# Compact but Moving: Intervention-Relevant Geometry in Recurrent World Models

Yuming Chen, Yang Liu
College of Intelligent Robotics and Advanced Manufacturing, Fudan University
Shanghai, China
{yumingchen, ly}@fudan.edu.cn

## Abstract:

Learned world models may have compact interventions even when their recurrent state is high-dimensional, but it is unclear what happens to such a correction after it enters the model. We study this question in a controlled recurrent world model where prior work identified a checkpoint-specific rank-4 interface for one-shot counterfactual velocity interventions. The correction rapidly leaves this fixed entry subspace during autonomous rollout. Nevertheless, a low-rank image obtained by transporting the entry directions through the factual recurrent Jacobian chain continues to capture most of the nonlinear correction. Restarts using the tangent-predicted correction preserve substantial counterfactual future function. This transport/function pattern recurs across independently trained structured-GRU models and a parameter-matched LSTM initialized with a privileged compact correction. We further characterize a finite-horizon future-response operator over the full recurrent carrier. Patching shifts its leading future-sensitive directions toward the matched native-counterfactual organization, and the local operator accurately ranks finite perturbation effects over the registered direction panels at the patched and native-counterfactual basepoints. A separate full-amplitude assay finds substantial factual-endpoint tangent residuals and supports response reconfiguration in two of three checkpoints. Together, these results show that compact intervention structure can persist as a moving, state-dependent local geometry embedded in high-dimensional recurrent dynamics, without implying a fixed or dynamically closed low-dimensional state.

Source codes and data can be found at https://github.com/lysea8282/future-response-dynamics

## 1 Introduction

World models compress observation histories into latent states that support prediction, planning, and imagined interaction (Ha & Schmidhuber, 2018; Hafner et al., 2020; Hafner et al., 2019; Hafner et al., 2021; Hafner et al., 2025; Schrittwieser et al., 2020). Yet predictive or control performance does not reveal how future-relevant information is organized inside those states. Intervention-based studies increasingly address this gap by asking which internal changes causally alter model behavior rather than merely what information can be decoded (Canby et al., 2025; Geiger et al., 2022; Geiger et al., 2024; Joseph et al., 2026). The unresolved dynamical question is what happens after such an intervention enters a recurrent model: how is the induced correction carried forward by the model's own dynamics?

Prior work (Liu & Chen, 2026), hereafter Paper 1, provides a controlled starting point. In a recurrent world model with a 192-dimensional carrier, a checkpoint-specific rank-4 interface was the smallest tested interface that could map a factual state and bounded velocity request to a one-shot hidden-state patch and redirect a 12-transition autonomous counterfactual rollout. The result established compact intervention access, but not a compact recurrent state. It left open whether the correction remains in the original low-rank plane, whether compact structure persists downstream in a state-dependent form, and how that structure relates to the carrier directions that affect future behavior.

Here we follow the correction after entry. Factual Jacobian transport shows that the correction rapidly leaves the fixed rank-4 entry plane yet remains concentrated in a trajectory-dependent low-rank image; tangent restart shows that this transported structure retains substantial counterfactual future function. The transport/function pattern persists across independently trained structured-GRU models and a parameter-matched monolithic LSTM comparator. We then characterize a finite-horizon future-response operator over the full carrier. Patching shifts its leading future-sensitive directions toward the matched native-counterfactual organization, and the operator accurately ranks finite effects over the registered perturbation panel. A separate factual-to-patched finite-dose assay shows that one factual-endpoint tangent is insufficient at the full patch amplitude and that the local response map reconfigures along the path.

These results separate several notions that are easily conflated. A compact intervention interface can produce a moving, trajectory-dependent compact structure rather than remain confined to the entry subspace; that transported structure can remain behaviorally effective while embedded in the full recurrent computation; and locally predictive response geometry need not define one global linear description at finite scale. Compactness in a recurrent world model can therefore reside in moving local geometry without implying a fixed or dynamically closed low-dimensional state.

## 2 Related work

### 2.1 Prior work and the downstream question

Intervention-based analyses increasingly distinguish information that is merely decodable from internal structure that can be changed causally. Distributed causal alignments can occupy subspaces rather than individual units, and intervention quality must be assessed by both behavioral effect and selectivity (Canby et al., 2025; Geiger et al., 2024). Recent work has begun to probe and steer physical variables in learned world and video models (Alam, 2026; Joseph et al., 2026; Zhang, 2026), but an immediate output change does not show that recurrent dynamics can sustain the altered future.

Paper 1 (Liu & Chen, 2026) addressed this stronger criterion in the same controlled two-object world used here. A checkpoint-specific rank-4 interface was the smallest tested interface that could map a factual state and a bounded velocity request to a one-shot hidden-state patch and then redirect a 12-transition autonomous rollout. The procedure replicated across fresh checkpoints and remained usable at nearby anchors. Paper 1 therefore established a compact dynamics-effective intervention-entry interface, not a four-dimensional closed recurrent state. It left open what happens after entry: whether compact structure persists downstream, how the correction is transported through the full carrier, which local carrier directions matter for a specified future, and where a local linear description breaks down. Those are the questions studied here.

### 2.2 Learned state and causal intervention

World models use latent states to summarize past observations and predict or simulate future outcomes (Ha & Schmidhuber, 2018; Hafner et al., 2020; Hafner et al., 2019; Hafner et al., 2021; Hafner et al., 2025; Schrittwieser et al., 2020). Classical POMDP belief states and predictive state representations make a stronger requirement explicit: a useful state must preserve the information needed for relevant futures (Boots et al., 2011; Kaelbling et al., 1998; Littman et al., 2001; Singh et al., 2004). Bisimulation and latent-state abstraction similarly ask which distinctions can be removed without changing behaviorally relevant transitions or rewards (Ferns et al., 2011; Gelada et al., 2019; Zhang et al., 2021). These notions motivate predictive sufficiency and closure, but they do not follow merely from low latent dimension.

Representation-intervention work asks a complementary causal question: which internal directions are actually used by the model. Causal abstraction, interchange interventions, and distributed alignment show that behaviorally meaningful variables may be distributed across learned subspaces rather than isolated coordinates (Geiger et al., 2021; Geiger et al., 2022; Geiger et al., 2024). Probe controls, concept-sensitivity measures, and amnesic or steering interventions distinguish decodable information, local sensitivity, and behavioral effects (Canby et al., 2025; Elazar et al., 2021; Hewitt & Liang, 2019; Kim et al., 2018; Rimsky et al., 2024; Stoehr et al., 2024). Recent world-model studies extend these ideas to physical and environmental variables (Alam, 2026; Joseph et al., 2026; Zhang, 2026). Our focus is the recurrent consequence of such an intervention: whether a compact correction remains organized and behaviorally effective as the model evolves autonomously.

This distinction is central to the present paper. A low-rank intervention interface, a low-rank transported correction, a low-rank future-sensitive subspace, and a closed effective state answer different questions. We therefore treat predictive sufficiency and closure as separate properties to be tested, rather than inferred from a single rank measured in hidden space (Sakcak et al., 2024).

### 2.3 Tangent dynamics and reduced descriptions

Local linearization provides the natural tool for following a perturbation through nonlinear dynamics. Along a reference trajectory, products of recurrent Jacobians transport infinitesimal perturbations across time (Dieci & Van Vleck, 2002; Ginelli et al., 2007; Hirsch et al., 2013). When only a low-dimensional direction set is followed, the relevant object is a moving subspace; principal angles and Grassmann geometry provide coordinate-free ways to compare such subspaces as their orientation changes (Edelman et al., 1998; Golub & Van Loan, 2013; Stewart & Sun, 1990).

Model-reduction methods provide a related but stronger objective: replacing high-dimensional dynamics with a smaller system while preserving selected behavior. Balanced truncation, POD/DMD, and Koopman-based approaches identify reduced coordinates or modes under different dynamical criteria (Antoulas, 2005; Korda & Mezić, 2018; Lusch et al., 2018; Moore, 1981; Otto & Rowley, 2019; Rowley, 2005; Schmid, 2010; Willcox & Peraire, 2002). Here we do not fit a reduced dynamics model. We keep the trained recurrent carrier fixed and use tangent operators to ask whether an intervention-induced compact structure remains locally compact as it is transported, and how that local geometry relates to future behavior.

## 3 Method

### 3.1 Experimental setting and inherited intervention interface

We use the controlled two-object environment and deterministic structured-GRU world model from Paper 1. Each object contributes position and velocity to an eight-coordinate primitive state; the complete recurrent carrier has 192 dimensions. The intervention anchor is $t = 7$: frames 0–4 and the anchor are observed, and frames 5–6 are masked. After the anchor, the model rolls out for 12 autonomous transitions with zero actions and no future observations. Single and Joint edits change one or two velocity components of the same object. Factual (F) and counterfactual (CF) histories share pre-anchor observations and realized observation noise. The native-counterfactual route assimilates the edited anchor observation through the ordinary model update. S1 contains units with no contact in the evaluated future; in S2, the edit changes future contact behavior. Appendix A gives the full specification.

Paper 1 established a checkpoint-specific rank-4 operational interface for one-shot velocity interventions. Given the factual primitive anchor state and the requested edit, a frozen affine map predicts four coefficients and applies

$$\hat{\mathbf{c}} = \mathbf{W}_4^{\mathrm{T}}\tilde{\mathbf{x}} + \mathbf{b}_4; \qquad \mathbf{z}_t^P = \mathbf{z}_t^F + \mathbf{U}_4\hat{\mathbf{c}} \tag{1}$$

Here, $\hat{\mathbf{c}} \in \mathbb{R}^4$ contains the predicted coefficients, $\mathbf{U}_4 \in \mathbb{R}^{192\times4}$ is the orthonormal entry basis, and $\mathbf{W}_4$ and $\mathbf{b}_4$ define the frozen affine map. Its input $\tilde{\mathbf{x}} \in \mathbb{R}^{12}$ concatenates the requested velocity edit and factual primitive anchor state, standardized with interface-fit statistics. Superscripts F and P denote factual and once-patched carriers. The interface is fitted on 1024 designated units and does not access the evaluation unit's native-counterfactual carrier. No further patch is applied after release (Appendix A).

### 3.2 Factual tangent transport of the intervention-induced correction

Let $k = 0$ denote the post-assimilation anchor and $\mathbf{z}_{t,k}^{\mathrm{q}}$ the complete carrier after $k$ releases on route $q$. The realized correction is $\boldsymbol{\delta}_k = \mathbf{z}_{t,k}^{\mathrm{P}} - \mathbf{z}_{t,k}^{\mathrm{F}}$; $\boldsymbol{\delta}_0$ is its anchor value. For the frozen autonomous transition $T$ and zero action $\mathbf{a}_{t+k}$, the factual full-carrier Jacobian is $\mathbf{J}_k^{\mathrm{F}} = D_{\mathbf{z}}\mathcal{T}(\mathbf{z}_{t,k}^{\mathrm{F}}, \mathbf{a}_{t+k})$. Its ordered product defines the tangent propagator:

$$\boldsymbol{\Phi}_{0\leftarrow0}^{\mathrm{F}} = \mathbf{I}_{192}; \qquad \boldsymbol{\Phi}_{k\leftarrow0}^{\mathrm{F}} = \mathbf{J}_{k-1}^{\mathrm{F}} \cdots \mathbf{J}_0^{\mathrm{F}}, \quad k \geq 1 \tag{2}$$

We apply the same propagator to the realized anchor correction and to the inherited rank-4 entry basis:

$$\hat{\boldsymbol{\delta}}_k^{\mathrm{F}} = \boldsymbol{\Phi}_{k\leftarrow0}^{\mathrm{F}}\boldsymbol{\delta}_0; \qquad \mathbf{U}_{4,k}^{tr} = \boldsymbol{\Phi}_{k\leftarrow0}^{\mathrm{F}}\mathbf{U}_4; \qquad \mathbf{Q}_k = \mathrm{orth}(\mathbf{U}_{4,k}^{tr}) \tag{3}$$

The first quantity is the tangent-predicted correction; the second is the raw transported image of the entry directions. Orthonormalization is used only to represent the image's numerical column space. We measure how much of the actual later patched-minus-factual correction remains in the original entry plane and how much is captured by the transported image:

$$\mathrm{OccFixed}_k = \frac{\|\mathbf{U}_4{\mathbf{U}_4}^{\mathrm{T}}\boldsymbol{\delta}_k\|^2}{\|\boldsymbol{\delta}_k\|^2 + \varepsilon}; \quad \mathrm{Cap}_k = \frac{\|\mathbf{Q}_k{\mathbf{Q}_k}^{\mathrm{T}}\boldsymbol{\delta}_k\|^2}{\|\boldsymbol{\delta}_k\|^2 + \varepsilon} \tag{4}$$

Here, the denominators use $\varepsilon = 10^{-24}$. Because the transported image is constructed along each unit's factual trajectory, its orientation can change with state and release time. Raw basis columns are propagated through the Jacobian chain; orthonormalization is used only for capture measurements. Figure 1 covers releases 1–12; transport/function confirmation uses releases 1–10 (details can be found in Appendix B).

To test function, we restart the unchanged model from $\mathbf{z}_{t,k}^T = \mathbf{z}_{t,k}^F + \hat{\boldsymbol{\delta}}_k^F$ and release it autonomously for the remaining future. Recovery measures the fraction of the patched route's reduction in normalized error to native CF that is retained by this restart. It is evaluated only where the patched improvement over factual error exceeds $10^{-5}$; these are the contrastable unit-times. Appendix B.5 gives the error and control definitions.

To test transport under a different recurrent organization, we use a parameter-matched monolithic single-layer LSTM (494664 parameters versus 496200). Its complete restart/differentiation carrier combines a 192-dimensional hidden state and a 192-dimensional cell state, giving 384 dimensions; only the hidden state feeds the decoder. Development selected rank 6 as the smallest common tested capacity rank. Fresh checkpoints 392001–392003 use a privileged anchor correction obtained by projecting their native-CF-minus-factual carrier difference onto a checkpoint-specific rank-6 basis. No later native-CF carrier enters tangent transport. Architecture and calibration are in Appendix A.5; the correction and schematic are in Appendix B.6 and Figure B1.

### 3.3 Finite-horizon full-carrier future-response operator

Transport follows one correction through time. The separate structured-GRU response assay asks which full-carrier directions influence a specified future. For $q \in \{\mathrm{F}, \mathrm{P}, \mathrm{CF}\}$, let $G_{t,H}$ stack the eight decoded physical coordinates over the next $H = 8$ autonomous transitions, using the frozen coordinate scaling in Appendix C.1. Its Jacobian at $\mathbf{z}_t^{\mathrm{q}}$ is

$$\mathbf{R}_{t,H}^{(q)} = \left.\frac{\partial G_{t,H}(\mathbf{z})}{\partial \mathbf{z}}\right|_{\mathbf{z}=\mathbf{z}_t^{(q)}} \in \mathbb{R}^{64\times 192} \tag{5}$$

The operator maps 192 carrier coordinates to the 64-dimensional stacked future query. Applied to a sufficiently small carrier perturbation, it gives the first-order future-output change. We evaluate it at matched factual, patched, and native-counterfactual anchors while holding checkpoint, future actions, horizon, decoder, and output scaling fixed. Unlike a one-step recurrent Jacobian, it includes recurrent propagation and decoding across the entire eight-transition query. Appendix C gives the full map and chain-rule construction.

### 3.4 Future-sensitive subspaces and CF-directed alignment

Right singular vectors of the response operator identify carrier directions with high future sensitivity. For each unit, $r_{\mathrm{CF}}$ is the smallest native-CF rank accounting for at least 90% of the squared singular-value sum. The same rank is used for all matched route comparisons. We write $\mathrm{S}_{\mathrm{sub}}$ for the mean squared cosine of principal angles between the corresponding leading right-singular subspaces; one denotes coincident subspaces (Appendix D).

The primary unit-level contrast asks whether patching moves the leading future-sensitive subspace toward the matched native-counterfactual subspace:

$$\Delta_{\mathrm{Align}} = S_{\mathrm{sub}}^{\mathrm{P,CF};r_{\mathrm{CF}}} - S_{\mathrm{sub}}^{\mathrm{F,CF};r_{\mathrm{CF}}} \tag{6}$$

Positive values indicate greater patched-to-native-CF than factual-to-native-CF overlap under the same comparison rank. Direct patched-factual subspace separation and full-operator Frobenius comparisons are descriptive/supporting only; Appendix D gives their definitions.

### 3.5 Local future-response sensitivity and perturbation validation

Subspace alignment does not by itself show that larger local response gains produce larger realized rollout effects. We therefore evaluate a frozen 32-unit subset (16 S1 and 16 S2 per checkpoint) at both patched and native-counterfactual basepoints. Each route uses 52 unit-norm directions generated from the same registered six-family specification and ordering; coordinate vectors are route-specific except for four shared Paper 1 interface directions. Some probes are operator-derived, and the projected-random family is conditioned on the local response subspace. Appendix E gives the complete panel and deterministic seed construction.

For a registered unit direction, the predicted effect is the norm of its response under the local operator. Finite effects are measured by perturbing that same route-specific vector in both signs with $\epsilon =$ $0.10\|\mathbf{z}_t^{\mathrm{P}} - \mathbf{z}_t^{\mathrm{F}}\|_2$, then releasing the unchanged model. The two future-deviation norms are added and divided by $2\epsilon$. Let $\mathbf{g}_{\mathrm{pred}}^{q}$ and $\mathbf{g}_{\mathrm{actual}}^{q}$ collect the 52 predicted and measured effects. Their unit-level Spearman correlation is

$$\rho^{(q)} = \mathrm{Corr}\left(\mathrm{rank}_{\mathrm{avg}}\left(\mathbf{g}_{\mathrm{pred}}^{(q)}\right), \mathrm{rank}_{\mathrm{avg}}\left(\mathbf{g}_{\mathrm{actual}}^{(q)}\right)\right) \tag{7}$$

This tests whether the local operator ranks finite autonomous effects over the registered perturbation panel. It does not test calibrated effect magnitudes, arbitrary unseen directions, or paired P-versus-CF effects on identical coordinate vectors. Patched and native-counterfactual routes are scored separately before aggregation.

### 3.6 Finite-dose response reconfiguration and path accounting

The perturbation assay above is local around the patched and native-counterfactual basepoints. We separately test how the response map behaves over the full inherited factual-to-patched carrier displacement. For each matched unit, the hidden-dose path is

$$\Delta\mathbf{z}_t^{(P)} = \mathbf{z}_t^{(P)} - \mathbf{z}_t^{(F)}; \qquad \mathbf{z}_\gamma = \mathbf{z}_t^{(F)} + \gamma\Delta\mathbf{z}_t^{(P)}; \qquad 0 \le \gamma \le 1 \tag{8}$$

The path joins the factual and full patched anchor carriers; it is not a physical-edit dose axis or an interpolation toward native CF. Let $\Delta\mathbf{Y}_{t,H}^{(PF)} = G_{t,H}\left(\mathbf{z}_t^{(P)}\right) - G_{t,H}\left(\mathbf{z}_t^{(F)}\right) \in \mathbb{R}^{64}$. The normalized error of predicting this endpoint future change from the factual tangent is

$$E_{\mathrm{local}} = \frac{\left\|\Delta\mathbf{Y}_{t,H}^{(PF)} - \mathbf{R}_{t,H}^{(F)}\Delta\mathbf{z}_t^{(P)}\right\|_2}{\left\|\Delta\mathbf{Y}_{t,H}^{(PF)}\right\|_2 + 10^{-12}} \tag{9}$$

Let $\mathbf{R}_\gamma$ denote the operator in Eq. (5) evaluated at $\mathbf{z}_\gamma$, so that $\mathbf{R}_0 = \mathbf{R}_{t,H}^{\mathrm{F}}$ and $\mathbf{R}_1 = \mathbf{R}_{t,H}^{\mathrm{P}}$. The registered reconfiguration statistic measures the midpoint operator's departure from the straight endpoint chord:

$$C_R(0.5) = \frac{\|\mathbf{R}_{0.5} - 1/2\,(\mathbf{R}_0 + \mathbf{R}_1)\|_F}{\|\mathbf{R}_1 - \mathbf{R}_0\|_F + 10^{-12}} \tag{10}$$

The midpoint quantity is the only registered operator-reconfiguration gate. A 33-node path with nested composite-Simpson rules is used only as numerical accounting to verify that integrated local directional responses recover the finite endpoint effect. Appendix F gives the complete path, quadrature, and numerical-validity definitions.

### 3.7 Populations, aggregation, and decision rules

The future-response assay uses frozen checkpoints 291402–291404. Alignment and finite-dose analyses use 128 matched units per checkpoint (64 S1 and 64 S2); perturbation validation uses a frozen 32-unit subset (16 per stratum) at both patched and native-counterfactual base routes. Unit identities, strata, direction-construction rules, and dose nodes are fixed before the corresponding effects are evaluated.

Aggregation is checkpoint first: units, directions and dose nodes refine within-checkpoint summaries rather than create additional replications. For the three-checkpoint response panel, replication requires support in at least two checkpoints under all required strata and routes. Alignment requires positive median CF-directed gain in both contrastable strata and positive equal-checkpoint family medians. Perturbation predictiveness requires median unit-level Spearman correlation greater than 0.70 separately in each checkpoint–stratum–route cell. Finite-dose support requires both effect medians greater than 0.10 in each stratum, together with the registered numerical-validity checks.

Structured-GRU tangent transport uses its separately frozen checkpoint panels. The LSTM comparison uses fresh checkpoints 392001–392003 and the development-frozen rank-6 privileged correction; support requires the registered capture, recovery, contrastability, tangent-versus-factual, and numerical-validity criteria in both strata. All gates are conjunctive and fixed before outcome inspection. Appendices B.5–B.6 and G give the assay definitions, decision rules, and provenance.

## 4 Results

### 4.1 The compact correction leaves its fixed entry subspace but remains captured by its transported image

Figure 1 shows that the one-shot correction leaves the fixed rank-4 entry plane almost immediately while remaining concentrated in its trajectory-specific transported image. Across checkpoints 291402–291404, fixed-entry occupancy at the first release was 0.8–13.7%, whereas transported-image capture was 86.9–97.4%; at the twelfth release, occupancy remained 1.4–10.6% while capture remained 87.8–98.3%.

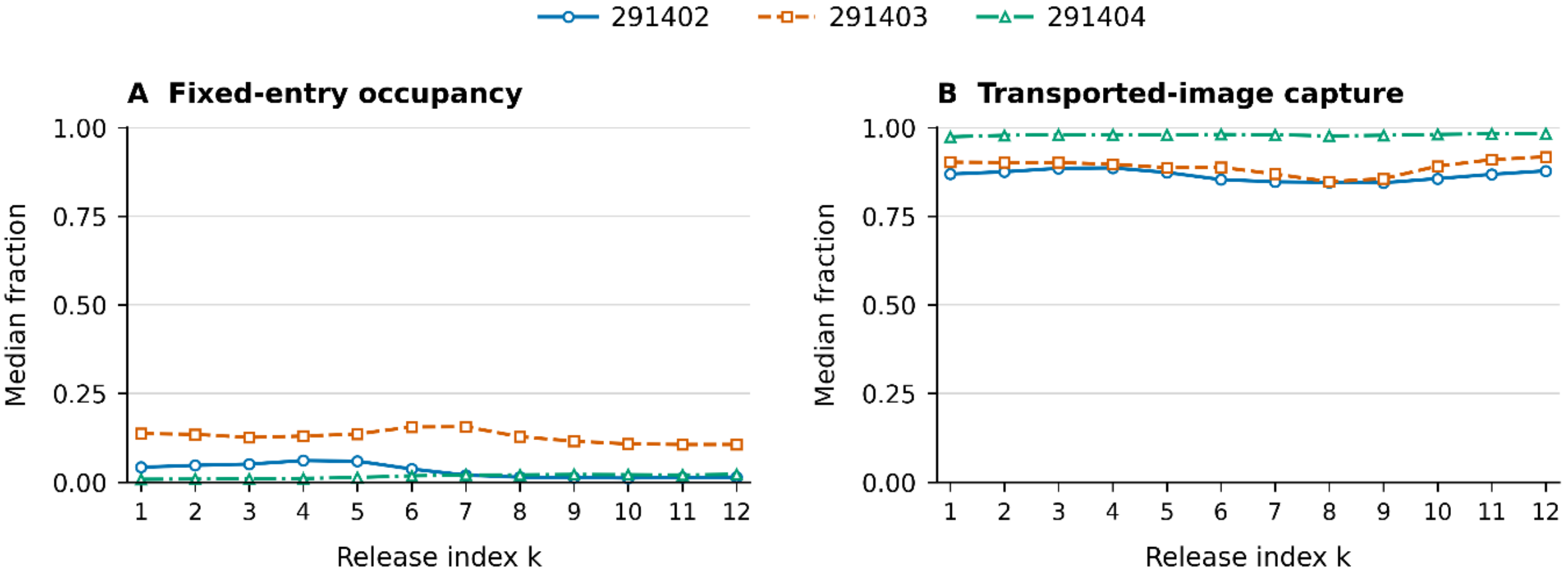

Figure 1. Correction geometry for checkpoints 291402–291404. Curves are within-checkpoint medians over the 1024-unit interface-fit population across releases k=1–12. (A) Squared correction fraction in the fixed entry plane. (B) Squared fraction in its factual-Jacobian-transported image. This historical geometry population is distinct from the held-out confirmation registry.

Transported-image capture also replicated in a separate confirmation cohort: all 18 checkpoint-by-stratum cells from nine baseline-eligible structured-GRU checkpoints passed, with median capture of 0.907–0.987. The original three curves summarize checkpoint-level results, while this confirmation uses separately reserved evaluation units. The evidence supports capture by trajectory-dependent transported images rather than continued occupancy of the entry plane.

### 4.2 Tangent-transported corrections retain substantial future function across trained models

The transported correction also retained future function. Across the nine baseline-eligible structured-GRU checkpoints, all 18 checkpoint-by-stratum cells satisfied the transport-capture and functional criteria; median normalized recovery ranged from 0.744 to 1.034, with 17/18 cells above 0.85. Recovery = 0 corresponds to the factual route, recovery = 1 to the operational patched route, and values above 1 indicate a tangent restart closer to native CF than the patched route. Thus, higher recovery indicates better restoration of the counterfactual future, with a value of 1 matching the improvement achieved by the original patch (as shown in Figure 2).

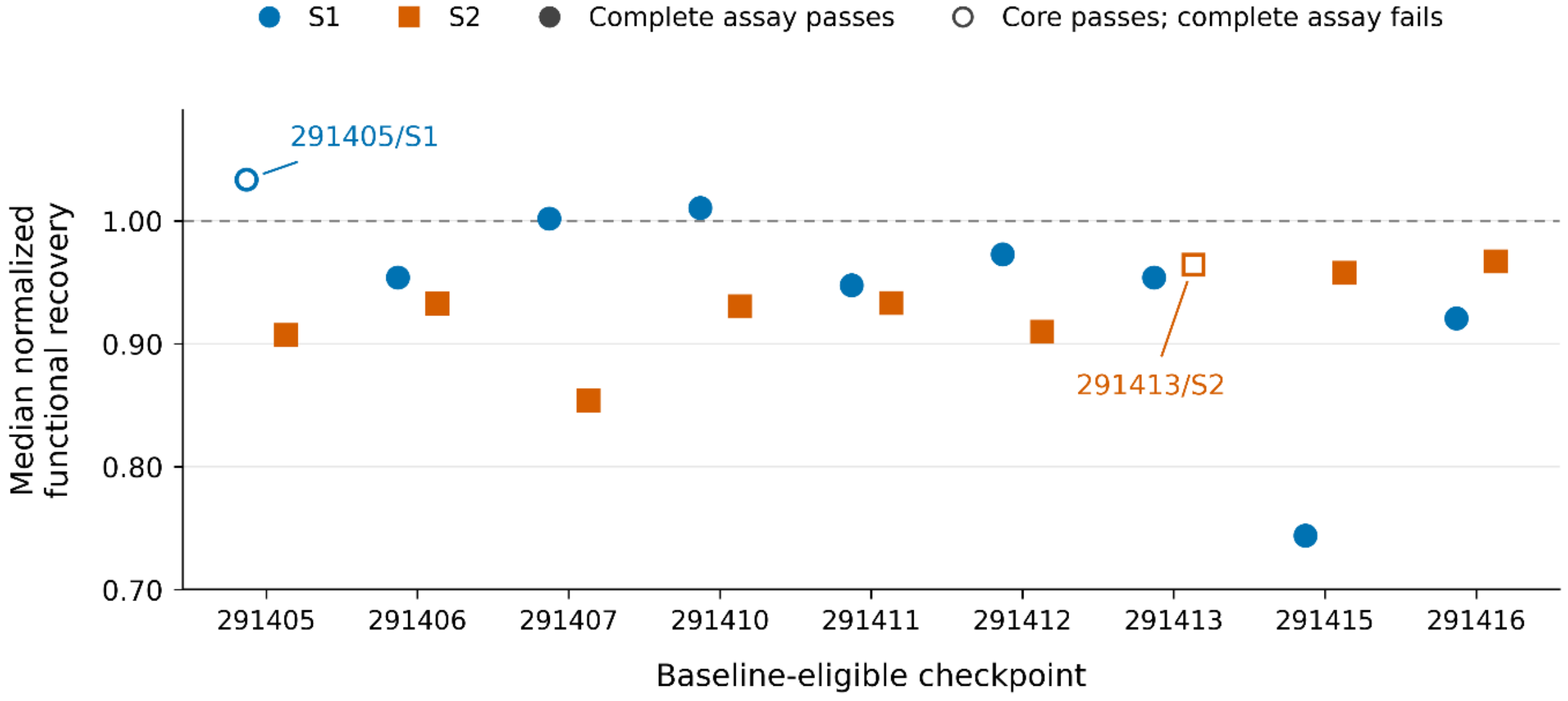

Figure 2. Tangent-restart recovery from the fixed 64-unit confirmation registry, summarized by checkpoint and stratum over releases 1–10. S1 and S2 are shown separately. Filled markers pass the complete assay; open markers retain transport/function support but fail contrastability/specificity requirements. The line at recovery = 1 denotes the operational patched route, not a pass threshold. Checkpoint 291414 is omitted because it failed baseline health.

The stricter complete assay passed 16/18 eligible checkpoint-by-stratum cells and 7/10 checkpoints in the fixed sample (7/9 among baseline-eligible checkpoints). The two failing cells retained high capture and recovery but failed contrastability/specificity, showing that the transport/function core is more reproducible than full operational specificity. Exact checkpoint-level details are given in Appendix H.1.

### 4.3 The transport core generalizes to a monolithic LSTM

All three fresh LSTM checkpoints supported the transport/function assay in both S1 and S2. Across the six checkpoint-by-stratum cells, transported-image capture was 0.936–0.997 and normalized recovery was 0.894–0.986; positive-unit fraction was 1.0 throughout and contrastable coverage was 0.95–1.0. Late-band capture remained 0.866–0.996 and recovery 0.809–0.977. Exact checkpoint values are reported in Appendix H.2.

The LSTM used a different recurrent structure and state dimension, with a rank-6 correction selected in development tests. It reproduced the state-dependent transport of compact corrections observed in the GRU, showing that this behavior extends across the two architectures.

### 4.4 Patching shifts future-sensitive subspaces toward the native counterfactual

Patching also changed the local future-sensitive organization. The CF-directed alignment gain from Section 3.4 was positive in all six checkpoint-by-stratum cells, so all three response-assay checkpoints supported the component in both S1 and S2. Checkpoints 291402 and 291403 showed gains of 0.082–0.098, whereas checkpoint 291404 showed much smaller positive gains (0.0043 in S1 and 0.0063 in S2), revealing substantial checkpoint heterogeneity.

The supporting full-operator comparison was positive in five of six cells and slightly negative in 291404/S1. The leading future-sensitive carrier subspaces showed a reproducible shift in local orientation toward their matched native-CF counterparts. Convergence of the full response operator was less consistent. Exact subspace and descriptive readbacks are in Appendix H.3.

### 4.5 Local future-response sensitivity predicts the rank ordering of finite perturbation effects

The local future-response operator also predicted the ordering of finite effects over the registered perturbation panels. Across all 12 checkpoint-by-stratum-by-route cells, median unit-level Spearman correlation was 0.9994–0.9999, above the registered 0.70 threshold; patched and native-counterfactual routes passed in both strata for all three checkpoints. Thus the perturbation-predictiveness component was supported by 3/3 checkpoints.

This result only ranks effects within the registered direction panel. It confirms finite autonomous effects on the tested structured and random directions, but does not compare calibrated magnitudes, unseen directions, or the two routes on identical vectors. Exact cell values are reported in Appendix H.4.

### 4.6 Finite-dose response reconfiguration replicates despite checkpoint heterogeneity

At the full inherited patch amplitude, the factual-endpoint linearization left a substantial residual. The median full-dose residual exceeded 0.10 in all six checkpoint-by-stratum cells (0.1888–1.1589), while the median midpoint operator-chord deviation exceeded 0.10 in five of six cells (0.0796–0.5202). Checkpoints 291402 and 291403 passed both strata; checkpoint 291404 failed S1 only because its midpoint deviation was 0.0796. The finite-dose component therefore replicated in 2/3 checkpoints.

Numerical path accounting was stable in all six cells, and both registered finite-dose quantities were larger in S2 than S1 for every checkpoint. Together with Section 4.5, these results show that locally predictive response geometry can reconfigure at finite scale: route-local operators rank effects accurately within their registered panels, while the separate factual-endpoint tangent does not describe the full patch with one fixed linear map. Exact finite-dose and numerical readbacks are in Appendix H.5.

## 5 Discussion

### 5.1 Compact entry becomes moving local geometry

Paper 1 established that bounded velocity edits can enter the structured-GRU through a checkpoint-specific rank-4 interface. The present results show that this compactness does not correspond to persistent storage in that fixed plane. The realized correction rapidly leaves the entry subspace, yet remains concentrated in the trajectory-dependent image obtained by factual tangent transport and retains substantial future function under tangent restart. Thus the intervention-relevant structure is compact but moving: the Jacobian chain provides a local first-order approximation to how the nonlinear correction is carried through the full recurrent state, rather than defining the hidden geometry itself.

This distinction separates intervention dimension from recurrent-state dimension. The rank-4 interface specifies how the tested counterfactual edits enter the model; it does not imply that four coordinates form a self-contained state. Conversely, leaving the original plane does not eliminate compact structure. The correction remains close to a low-rank image whose orientation changes with the factual trajectory. A complementary analysis shows that the patched state's leading future-sensitive directions shift toward those of the matched native-counterfactual state.

### 5.2 State-dependent transport and finite-scale response

The cross-architecture comparison shows which part of this picture is not specific to the structured-GRU. The monolithic LSTM uses a different recurrent organization, a 384-dimensional hidden-plus-cell carrier, and a development-frozen privileged rank-6 correction, yet its factual tangent dynamics likewise capture most later carrier displacement and preserve much of the correction's future function. The shared phenomenon is therefore state-dependent transport of a compact correction, not a common basis or common rank.

The future-response assays further show that local descriptions can be behaviorally predictive without remaining globally fixed. At the patched and native-counterfactual basepoints, the local operators almost perfectly rank finite effects over their registered perturbation panels. In the separate factual-to-patched finite-dose assay, however, one factual-endpoint tangent does not account for the complete future change, and the response operator reconfigures along the path. Integrating local directional responses along the fixed factual-to-patched anchor path numerically recovers the endpoint future change. This is a consistency check on finite-change accounting, separate from propagation over rollout time. Together, the assays support coherent correction transport and state-dependent future-response geometry.

### 5.3 Separating intervention geometry, response geometry, and effective state

These results separate several notions that are often conflated. The Paper 1 rank-4 result concerns a compact intervention interface; tangent transport concerns how an injected correction is carried; the future-response operator concerns which local carrier directions affect a specified future; and perturbation validation establishes behavioral relevance over the registered local panels. None of these quantities alone establishes dynamical closure.

A closed effective state would require a reduced representation that can update its own future-relevant information without repeatedly returning to the full recurrent carrier. The present experiments neither establish nor rule out such a representation. They instead show that compact intervention geometry and query-dependent future sensitivity coexist within the full recurrent computation. Closure therefore has to be tested directly rather than inferred from intervention rank, transport rank, or local response rank.

## 6 Conclusion

This work followed a compact counterfactual correction after it entered a recurrent world model. In the structured-GRU, the correction leaves its rank-4 entry plane but remains concentrated in a trajectory-dependent transported image. Tangent restarts retain substantial future function in that model and in an LSTM initialized with a privileged compact correction. Patching also reorganizes local future sensitivity, while finite-dose analysis shows that the factual-endpoint tangent does not account for the full patch.

Together, the results show that compactness can persist as moving local geometry embedded in state-dependent high-dimensional recurrent dynamics. This separates compact intervention, compact transport, local future sensitivity, behavioral relevance, and dynamical closure, and leaves the existence of a closed effective state as a distinct question.

## AI Assistance Disclosure

Generative AI tools, including ChatGPT and OpenAI Codex, were used as research-assistance tools for literature discovery and synthesis, refinement of research questions and experimental protocols, code implementation and debugging, analysis organization, interpretation of results and development of conceptual explanations, and manuscript preparation. All experimental protocols, scientific decisions, formal runs, interpretation of results, and final claims were reviewed and approved by the authors. AI-generated code, references, analyses, and text were reviewed by the authors and checked against the relevant source code, experimental records, and primary literature where applicable. The authors take full responsibility for the accuracy and integrity of the work.

## Appendix A. Environment, recurrent model, and inherited patch interface

### A.1 Simulator and observation interface

The controlled simulator contains two circular objects of radius 0.25 moving in a square reflecting arena. The center coordinate of each object is restricted to $[-4.75, 4.75]$ on each axis, and the integration time step is 0.25. For completeness, the joint primitive state is

$$\mathbf{s}_t = \left(x_{0,t}, y_{0,t}, v_{x,0,t}, v_{y,0,t}, x_{1,t}, y_{1,t}, v_{x,1,t}, v_{y,1,t}\right)^{\mathsf{T}} \in \mathbb{R}^8 \tag{A1}$$

Here, $\mathbf{s}_t$ is the eight-dimensional joint primitive state at time $t$; $i \in \{0,1\}$ indexes the object; $x_{i,t}$ and $y_{i,t}$ are its horizontal and vertical positions; and $v_{x,i,t}$ and $v_{y,i,t}$ are its horizontal and vertical velocity components.

The physical simulator is deterministic and contains no process noise. Within a time step, circle–circle collision detection solves analytically for the earliest contact time from the relative-motion quadratic. The simulator advances to contact, applies the implemented normal-impulse velocity update, and then advances through the remaining part of the step. Wall crossings are reflected by reversing the corresponding velocity component.

The observation at each time has shape $2 \times 6$. For object $i$, $\mathbf{o}_{i,t} \in \mathbb{R}^6$ denotes the observation vector containing normalized noisy $x_{i,t}$, $y_{i,t}$, $v_{x,i,t}$, $v_{y,i,t}$, a binary visibility indicator, and a fixed identity value. The identities for objects 0 and 1 are $-1$ and $+1$, respectively. When an object is visible, its position, velocity, and identity are provided directly. Radius and contact are not observation channels. Position and velocity observation noise have standard deviations 0.005 and 0.01, respectively. Frames 0–4 and the selected anchor are visible, while the physical observation channels after frame 4 and before the anchor are zeroed. For the main anchor $t = 7$, frames 5 and 6 therefore form a two-step unobserved interval before the anchor observation is assimilated. Factual and matched counterfactual branches use the same realization of observation noise.

The eight physical coordinates are normalized with the frozen mean vector

$$\boldsymbol{\mu}_s = \begin{pmatrix} 0.01091056,\ 0.00882761,\ -0.00132395,\ 0.00020721, \\ 0.01091056,\ 0.00882761,\ -0.00132395,\ 0.00020721 \end{pmatrix}^{\mathsf{T}} \tag{A2}$$

and the frozen standard-deviation vector

$$\boldsymbol{\sigma}_s = \begin{pmatrix} 0.67486298,\ 0.67908412,\ 0.29360580,\ 0.29595292, \\ 0.67486298,\ 0.67908412,\ 0.29360580,\ 0.29595292 \end{pmatrix}^{\mathsf{T}} \tag{A3}$$

Here, $\boldsymbol{\mu}_s \in \mathbb{R}^8$ and $\boldsymbol{\sigma}_s \in \mathbb{R}^8$ are the coordinatewise normalization mean and standard-deviation vectors in the primitive-state ordering of Eq. (A1). These values are fixed by the model/data interface.

### A.2 Recurrent model and carrier assembly

The deterministic structured-GRU world model contains 496200 trainable parameters. A shared observation encoder maps each six-dimensional object observation through $6 \rightarrow 128 \rightarrow 64$, and the action encoder maps the four-dimensional action input through $4 \rightarrow 128 \rightarrow 64$. The message network has dimensions $192 \rightarrow 256 \rightarrow 128 \rightarrow 64$. Object-local recurrence uses a shared GRU with 144-dimensional data input and 48-dimensional recurrent state, while the global GRU uses a 192-dimensional data input and a 64-dimensional recurrent state. The prior and posterior middle-component networks both have dimensions $176 \rightarrow 256 \rightarrow 128 \rightarrow 16$, and the decoder maps the complete carrier through $192 \rightarrow 384 \rightarrow 192 \rightarrow 8$. No normalization layers are used. All experiments considered here use zero actions. The structured-GRU RSSM architecture itself is unchanged from Paper 1, so we refer readers to Paper 1 for the corresponding schematic. For the monolithic LSTM comparator schematic, see Appendix B, Figure B1 below.

Let $\mathbf{h}_{i,t} \in \mathbb{R}^{48}$ denote the recurrent state of object $i$, $\mathbf{m}_{i,t} \in \mathbb{R}^{16}$ its middle component, and $\boldsymbol{gs}_t \in \mathbb{R}^{64}$ the global recurrent state. The object GRU receives the previous middle component, the object-interaction message, and the object-specific action embedding. The updated object states enter the global recurrent update. A prior middle component is then predicted from the updated object state, updated global state, and action embedding, whereas a posterior middle component is computed from the updated object state, updated global state, and encoded observation. Visibility selects between the two components,

$$\mathbf{m'}_{i,t} = v_{i,t}\mathbf{m}_{i,t}^{\text{post}} + \left(1 - v_{i,t}\right)\mathbf{m}_{i,t}^{\text{prior}} \tag{A4}$$

Here, $v_{i,t} \in \{0,1\}$ is the visibility indicator of object $i$ at time $t$; $\mathbf{m}_{i,t}^{\text{post}} \in \mathbb{R}^{16}$ and $\mathbf{m}_{i,t}^{\text{prior}} \in \mathbb{R}^{16}$ are its posterior and prior middle components; and $\mathbf{m'}_{i,t} \in \mathbb{R}^{16}$ is the selected middle component after visibility gating.

The complete carrier is packed as

$$\mathbf{z}_t = [\boldsymbol{h}'_{0,t}, \boldsymbol{m}'_{0,t}, \boldsymbol{h}'_{1,t}, \boldsymbol{m}'_{1,t}, \boldsymbol{gs}'_t] \in \mathbb{R}^{192}. \tag{A5}$$

Here, $\mathbf{z}_t$ is the complete 192-dimensional packed carrier; $\mathbf{h'}_{0,t}$ and $\mathbf{h'}_{1,t}$ are the updated 48-dimensional object recurrent states; $\mathbf{m'}_{0,t}$ and $\mathbf{m'}_{1,t}$ are the selected 16-dimensional middle components; and $\boldsymbol{gs}'_t$ is the updated 64-dimensional global recurrent state. The update is deterministic; there is no latent sampling or reparameterization step. The carrier stored at anchor $t$ is post-recurrent-update and post-observation-assimilation through that anchor.

The model-training split contains 4096 physical units with matched factual and local-counterfactual branches. The same physical training trajectories and realized observation noise are used across model seeds, while initialization and training order vary. The separately designated interface-fit population contains 1024 units and is disjoint from the held-out 64-unit transport/function registries. The historical geometry analysis in Figure 1 instead uses the 1024-unit interface-fit population itself.

Training uses 70 epoch equivalents, 24 batches per epoch, and 1680 optimizer updates. Each batch contains 96 physical units and 192 factual/counterfactual branches with equal branch weights. AdamW uses an initial learning rate of $1.5 \times 10^{-3}$, weight decay $10^{-5}$, and gradient-norm clipping at 5.0. A cosine schedule reduces the learning rate to $1.5 \times 10^{-4}$ over training. The terminal model after 1680 updates is used as the checkpoint.

### A.3 Checkpoint-specific rank-4 entry interface

The inherited intervention carrier is fitted independently for each checkpoint from $N = 1024$ Single carrier-fit units, where $N$ denotes the number of fit units. For fit unit $i$, define the native-counterfactual-minus-factual carrier difference at the common post-assimilation anchor as

$$\Delta\mathbf{z}_i = \mathbf{z}_{t,i}^{CF} - \mathbf{z}_{t,i}^{F} \in \mathbb{R}^{192} \tag{A6}$$

Here, $\Delta\mathbf{z}_i$ is the carrier difference for fit unit $i$; $\mathbf{z}_{t,i}^{CF}$ and $\mathbf{z}_{t,i}^{F}$ are its native-counterfactual and factual complete carriers at anchor $t$; and superscript $CF$ denotes the native-counterfactual route.

The raw differences are stacked row-wise,

$$\mathbf{D} = \begin{bmatrix} (\Delta \mathbf{z}_1)^\top \\ \vdots \\ (\Delta \mathbf{z}_N)^\top \end{bmatrix} \in \mathbb{R}^{N \times 192} \tag{A7}$$

where $\mathbf{D}$ is the uncentered carrier-difference matrix and $N = 1024$. It is factorized without centering,

$$\mathbf{D} = \mathbf{L}\mathbf{\Sigma}\mathbf{V}^\top \tag{A8}$$

Here, $\mathbf{L}$ and $\mathbf{V}$ contain the left and right singular vectors of $\mathbf{D}$, respectively, and $\mathbf{\Sigma}$ is the diagonal matrix of singular values.

If $\mathbf{v}_k$ denotes the $k$-th right singular vector, the rank-$r$ candidate carrier is

$$\mathbf{U}_r = [\mathbf{v}_1, \ldots, \mathbf{v}_r] \tag{A9}$$

Here, $r$ is the candidate rank, $\mathbf{v}_k \in \mathbb{R}^{192}$ is the $k$-th right singular vector, and $\mathbf{U}_r \in \mathbb{R}^{192 \times r}$ is the corresponding candidate entry basis. Its columns are orthonormal,

$$\mathbf{U}_r^\top \mathbf{U}_r = \mathbf{I}_r \tag{A10}$$

where $\mathbf{I}_r \in \mathbb{R}^{r \times r}$ is the identity matrix. Paper 1 selected $r = 4$ as the smallest tested rank that passed its registered intervention assay. The basis is fitted independently for each checkpoint; bases are not aligned across checkpoints.

### A.4 Addressable coefficient map and one-shot patch

For carrier-fit unit $i$, let $\mathbf{e}_i \in \mathbb{R}^4$ denote the requested velocity-edit vector in the slot order $(\Delta v_{x,0}, \Delta v_{y,0}, \Delta v_{x,1}, \Delta v_{y,1})$, and let $\mathbf{s}_{t,i}^F \in \mathbb{R}^8$ denote the factual primitive anchor state. The non-intercept affine input is

$$\mathbf{x}_i = \begin{bmatrix} \mathbf{e}_i \\ \mathbf{s}_{t,i}^F \end{bmatrix} \in \mathbb{R}^{12} \tag{A11}$$

Here, $\mathbf{x}_i$ is the 12-dimensional input vector for fit unit $i$. Single fitting uses exactly one nonzero component of $\mathbf{e}_i$.

Each non-intercept feature is standardized with carrier-fit statistics,

$$\tilde{x}_{i,k} = \frac{x_{i,k} - \mu_k}{\max(\sigma_k, 10^{-6})} \tag{A12}$$

Here, $x_{i,k}$ and $\tilde{x}_{i,k}$ are the raw and standardized values of feature $k \in \{1, \ldots, 12\}$ for unit $i$; $\mu_k$ and $\sigma_k$ are the carrier-fit mean and standard deviation of feature $k$; and $10^{-6}$ is the fixed scale floor. The resulting standardized vector is denoted $\tilde{\mathbf{x}}_i \in \mathbb{R}^{12}$.

The privileged fit target is the projection of the native carrier difference onto the checkpoint-specific rank-4 basis,

$$\mathbf{c}_i^{\text{oracle}} = \mathbf{U}_4^\top \Delta \mathbf{z}_i \tag{A13}$$

Here, $\mathbf{c}_i^{\text{oracle}} \in \mathbb{R}^4$ is the privileged coefficient target for fit unit $i$, $\mathbf{U}_4$ is the rank-4 basis from Eq. (A9), and $\Delta \mathbf{z}_i$ is defined in Eq. (A6).

The affine coefficient map is fitted by ridge regression,

$$(\mathbf{W}_4, \mathbf{b}_4) = \arg\min_{\mathbf{W},\mathbf{b}} \sum_{i=1}^{N} \left\| \mathbf{c}_i^{\text{oracle}} - \left(\mathbf{W}^\top \tilde{\mathbf{x}}_i + \mathbf{b}\right) \right\|_2^2 + 10^{-4} \parallel \mathbf{W} \parallel_F^2 ,$$

(A14)

Here, $\mathbf{W}_4 \in \mathbb{R}^{12\times4}$ and $\mathbf{b}_4 \in \mathbb{R}^4$ are the fitted weight matrix and intercept; $\mathbf{W}$ and $\mathbf{b}$ are the optimization variables; $N = 1024$ is the number of Single carrier-fit units; $\|\cdot\|_2$ is the Euclidean norm; and $\|\cdot\|_F$ is the Frobenius norm. The $10^{-4}$ penalty is applied to the weight matrix but not to the intercept. At evaluation time, the predicted coefficient vector is

$$\hat{\mathbf{c}} = \mathbf{W}_4^\top \tilde{\mathbf{x}} + \mathbf{b}_4$$

(A15)

where $\hat{\mathbf{c}} \in \mathbb{R}^4$ is the predicted rank-4 coefficient vector for the evaluation request and $\tilde{\mathbf{x}} \in \mathbb{R}^{12}$ is the corresponding standardized input. The one-shot patch is then

$$\mathbf{z}_t^P = \mathbf{z}_t^F + \mathbf{U}_4 \hat{\mathbf{c}}$$

(A16)

Here, $\mathbf{z}_t^F$ and $\mathbf{z}_t^P$ are the factual and once-patched complete carriers at anchor $t$, respectively. The implementation uses unit patch strength. The mapper receives only the factual primitive anchor state, the requested edit, and the frozen interface-fit statistics; it cannot access the evaluation unit's native-counterfactual carrier, future observations, simulator future, or oracle coefficients. The patch is applied once at the anchor, and the affine map is not consulted again during autonomous rollout. Paper 2 then uses the realized difference $\mathbf{z}_t^P - \mathbf{z}_t^F$ as the seed correction introduced in Section 3.1 and formalized by the matched-route definitions in Appendix B.

### A.5 Monolithic LSTM comparator architecture and calibration

To separate recurrent organization from parameter count, we use a monolithic deterministic single-layer LSTM comparator with 494664 trainable parameters, compared with 496200 in the structured-GRU reference. Each $2 \times 6$ observation is flattened to 12 scalars and encoded by a $12 \rightarrow 224 \rightarrow 224$ GELU MLP. The recurrent input concatenates the 224-dimensional encoded observation with the raw four-scalar action. A standard LSTM Cell with hidden size 192 produces hidden state $\mathbf{h}_t^L \in \mathbb{R}^{192}$ and cell state $\mathbf{c}_t^L \in \mathbb{R}^{192}$. The decoder maps $\mathbf{h}_t^L$ through $192 \rightarrow 256 \rightarrow 256 \rightarrow 8$ MLP layers with GELU activations after the two hidden layers and a linear output layer. Both recurrent states are initialized to zero, and the complete carrier used for restart and differentiation is $\mathbf{z}_t^L = [\mathbf{h}_t^L, \mathbf{c}_t^L] \in \mathbb{R}^{384}$. Figure B1 in Appendix B shows the comparator.

During autonomous rollout, the LSTM zeros the physical and visibility inputs and uses the fixed identity token $[0,1]$, whereas the observed history uses the dataset identity channels $[-1, +1]$. We report this implementation difference explicitly because unobserved-step inputs are not encoded identically across architectures.

Training was calibrated prospectively for the LSTM rather than forced to match the structured-GRU duration. Calibration used two development seeds and four predefined recipes. The selected C2 recipe uses 140 epochs and 3360 optimizer updates, with initial learning rate 0.0015 and final stored learning rate 0.00015, while retaining the inherited physical training data, objective, and terminal-checkpoint convention. Three fresh calibrated checkpoints, 392001, 392002, and 392003, passed baseline adequacy before the privileged transport experiment.

## Appendix B. Factual tangent transport and geometric capture

### B.1 Matched routes and autonomous release indexing

We use the complete-carrier notation from Appendix A and set $k=0$ at the common post-assimilation anchor $t=7$. For route $r\in\{\mathrm{F},\mathrm{P},\mathrm{CF}\}$, define

$$\mathbf{z}_{t,0}^{r}=\mathbf{z}_{t}^{r},\ \ \mathbf{z}_{t,k+1}^{r}=\mathcal{T}(\mathbf{z}_{t,k}^{r},\mathbf{a}_{t+k}) \tag{B1}$$

Here, $\mathbf{z}_{t,k}^{r}$ is the complete carrier after $k$ autonomous transitions from the anchor on route $r$; $\mathrm{F},\mathrm{P},\mathrm{CF}$ denote the factual, once-patched, and native-counterfactual routes; $\mathcal{T}$ is the frozen autonomous transition; and $\mathbf{a}_{t+k}$ is the recorded action. In all experiments, actions are set to zero. The rollout horizon is $H=12$, and all corresponding actions are zero. After the anchor, the model receives no future observation, teacher forcing, decoder feedback, repeated patch, physical clamp, or hidden-state clamp. The factual and patched routes have the same assimilated factual history through the anchor; the patched route differs only by the one-shot carrier change in Eq. (A16). The native-counterfactual route follows the same pre-anchor history and realized observation noise but assimilates the edited anchor observation through the ordinary model update. It is used later as the matched model-native future reference.

### B.2 Factual Jacobian chain

The one-step factual Jacobian at release index $k$ is

$$\mathbf{J}_{k}^{\mathrm{F}}=D_{\mathbf{z}}\mathcal{T}(\mathbf{z}_{t,k}^{\mathrm{F}},\mathbf{a}_{t+k}) \tag{B2}$$

Here, $D_{\mathbf{z}}$ denotes differentiation with respect to the complete carrier, so $\mathbf{J}_{k}^{\mathrm{F}}$ is a 192-by-192 Jacobian evaluated at the factual carrier for release index $k$.
The ordered product from the anchor to release $k$ is

$$\mathbf{\Phi}_{0\leftarrow0}^{\mathrm{F}}=\mathbf{I}_{192},\ \ \mathbf{\Phi}_{k\leftarrow0}^{\mathrm{F}}=\mathbf{J}_{k-1}^{\mathrm{F}}\mathbf{J}_{k-2}^{\mathrm{F}}\cdots\mathbf{J}_{0}^{\mathrm{F}} \tag{B3}$$

The multiplication order follows the temporal sequence from the anchor to the release point. $\mathbf{\Phi}_{k\leftarrow0}^{\mathrm{F}}$ acts on the full carrier and is evaluated entirely along the factual trajectory. The Jacobian-vector products and raw transported columns are propagated in the model's float32 arithmetic; no patched or native-counterfactual state is used to construct this chain.

### B.3 Transported correction and transported entry image

The tangent prediction of the anchor correction at release $k$ is

$$\widehat{\boldsymbol{\delta}}_{k}^{\mathrm{F}}=\mathbf{\Phi}_{k\leftarrow0}^{\mathrm{F}}\boldsymbol{\delta}_{0} \tag{B4}$$

Here, $\widehat{\boldsymbol{\delta}}_{k}^{\mathrm{F}}$ is predicted from the anchor correction and factual tangent dynamics only. The corresponding raw transported entry basis is

$$\mathbf{U}_{4,k}^{tr}=\mathbf{\Phi}_{k\leftarrow0}^{\mathrm{F}}\mathbf{U}_{4}\in\mathbb{R}^{192\times4} \tag{B5}$$

To obtain an orthonormal basis for its numerical column space, we compute the thin singular-value decomposition

$$\mathbf{U}_{4,k}^{tr}=\mathbf{L}_{k}\mathbf{\Sigma}_{k}\mathbf{R}_{k}^{\top} \tag{B6}$$

Here, $\mathbf{L}_{k}$ and $\mathbf{R}_{k}$ contain the left and right singular vectors of the transported basis, and $\mathbf{\Sigma}_{k}$ is the diagonal matrix of singular values. If $\sigma_{k,j}$ is the $j$-th singular value and $\sigma_{k,1}$ is the largest singular value, the retained numerical rank is

$$q_k = \mathrm{card}\{j\colon \sigma_{k,j} > \rho\sigma_{k,1}\},\ \ \rho = 2.288818359375 \times 10^{-5}$$
(B7)

The symbol $q_k$ denotes the retained numerical rank and $\rho$ is the frozen relative singular-value tolerance. The transported-image basis is

$$\mathbf{Q}_k = [\boldsymbol{\ell}_{k,1}, \dots, \boldsymbol{\ell}_{k,q_k}]$$
(B8)

Here, $\boldsymbol{\ell}_{k,j}$ is the j-th column of $\mathbf{L}_k$. Thus $\mathbf{Q}_k$ spans the numerically retained image of the original rank-4 entry basis after factual tangent transport. A loss of numerical rank is retained rather than replaced by an adaptively chosen higher-rank basis.

### B.4 Geometric measurements and release populations

The actual later patched-minus-factual correction is

$$\boldsymbol{\delta}_k = \mathbf{z}_{t,k}^{\mathrm{P}} - \mathbf{z}_{t,k}^{\mathrm{F}}$$
(B9)

Using $\varepsilon = 10^{-24}$ as the frozen numerical floor, fixed-entry occupancy and transported-image capture are

$$\mathrm{OccFixed}_k = \frac{\|\mathbf{U}_4\mathbf{U}_4{}^{\top}\boldsymbol{\delta}_k\|^2}{\|\boldsymbol{\delta}_k\|^2 + \varepsilon} \quad \mathrm{Cap}_k = \frac{\|\mathbf{Q}_k\mathbf{Q}_k{}^{\top}\boldsymbol{\delta}_k\|^2}{\|\boldsymbol{\delta}_k\|^2 + \varepsilon}$$
(B10)

The historical geometry analysis evaluates these quantities on the 1024-unit carrier-fit population over the full anchor-relative rollout. The fresh structured-GRU transport assay uses a separate 64-unit evaluation registry with 32 S1 and 32 S2 units and evaluates release indices $k = 1, \dots, 10$. S1 contains units with no contact event in the evaluated future, whereas S2 contains units for which the velocity edit changes future contact behavior. For descriptive time-course summaries, releases 1–3, 4–6, and 7–10 form the early, middle, and late bands, respectively.

Later patched states enter Eq. (B10) only to measure the realized correction. They are never supplied to the tangent construction in Eqs. (B2)–(B8). The model and Jacobian-vector chain use float32 arithmetic, while the thin SVD and projection measurements use float64. These calculations are performed in the native carrier coordinates of each checkpoint.

### B.5 Functional restart, recovery, and structured-GRU controls

Baseline eligibility was determined independently of transport outcomes on 256 historical units per stratum. For each checkpoint, factual (G0), native-counterfactual (G1), and full-hidden restart-equivalence (G2) health checks had to pass separately in S1 and S2. In G0 and G1, all units had to satisfy engineering and autonomy eligibility, at least 80% had to jointly pass all five frozen health metrics, and each metric's cell median had to satisfy the corresponding stratum-specific upper bound. Equality with a bound passed.

G2 required all 256 units to have both hidden-state and rollout maximum absolute restart discrepancies no larger than 10^-6. Checkpoint 291414 failed the S1/G1 joint-coverage requirement (192/256 = 0.75 < 0.80); it was therefore not evaluated for the mechanism, remained in the fixed ten-model end-to-end denominator, and was excluded from the nine-model baseline-eligible denominator. No checkpoint was replaced.

Table B1. Frozen structured-GRU baseline-health upper bounds used for G0/G1 eligibility.

| Metric | Definition | S1 upper bound | S2 upper bound |
|---|---|---|---|
| M1 | Standardized edited-velocity anchor error | 0.38875516163789126 | 0.4669759655361256 |
| M2 | RMS derived-state anchor error | 0.6530927312234215 | 0.6568131957898696 |
| M3 | Future primitive-state RMSE vs. simulator truth | 0.28744727060560543 | 0.5513987778513324 |
| M4 | Maximum unaffected-component anchor error | 0.3393832556356259 | 0.25616304846974386 |
| M5 | Post-edit law/boundary residual | 20.260185026563704 | 10.47115975990891 |

To test whether the tangent-predicted correction retains future function, we construct at release $k$ the synthetic tangent-restart carrier

$$\mathbf{z}_{t,k}^{T} = \mathbf{z}_{t,k}^{F} + \widehat{\boldsymbol{\delta}}_{k}^{F} \tag{B11}$$

The unchanged model is then released autonomously for the remaining future. No later observation, repeated projection, hidden-state clamp, physical clamp, native-counterfactual hidden state, or decoder feedback is supplied. Let $\mathbf{y}_j^{(R,k)} \in \mathbb{R}^8$ denote the physically denormalized prediction at future step $j$ from route $R$ after restart at release $k$, let $\mathbf{y}_j^{CF}$ denote the matched native-counterfactual decoded future, and let $\mathbf{s} \in \mathbb{R}^8$ be the frozen coordinate-wise error-scale vector. The remaining-future error is

$$e_R(u,k) = \sqrt{\frac{1}{8(H-k)} \sum_{j=k+1}^{H} \sum_{d=1}^{8} \left( \frac{y_{j,d}^{(R,k)} - y_{j,d}^{CF}}{s_d} \right)^2} \qquad H = 12 \tag{B12}$$

Here, $u$ indexes the registered physical unit. The current release frame is excluded, and all remaining future steps receive equal weight. Define

$$A(u,k) = e_F(u,k) - e_X(u,k) \qquad C(u,k) = \mathbf{1}[A(u,k) > 10^{-5}] \tag{B13}$$

and, only for contrastable unit-times with $C(u,k) = 1$,

$$\mathrm{Rec}(u,k) = \frac{e_F(u,k) - e_T(u,k)}{A(u,k)} \tag{B14}$$

For the structured-GRU assay, $X = P$ is the operational once-patched entry route; for the LSTM privileged assay below, $X = O$ is the privileged entry route. Recovery is not clipped. Values above one mean that the tangent restart is closer to the native-counterfactual future than the corresponding entry route under this error measure; they are not interpreted as a fraction of simulator-causal truth.

The structured-GRU confirmation additionally uses three matched controls: a sign-flipped anchor correction, a wrong-object request with the same axis and signed edit, and a deterministic in-plane sham direction orthogonal to the fitted coefficient vector. Each sham is norm-matched at release to the correct tangent-transported correction. These controls test route specificity for the inherited structured-GRU operational interface and are not transferred as primary semantic controls to the LSTM comparator.

The complete structured-GRU assay uses all ten registered releases in each checkpoint and stratum. Nested summaries first take the temporal median within each unit and then the median across units. Contrastable coverage is the fraction of the 32 × 10 registered unit-times satisfying the contrastability condition in Eq. (B13); positive-unit coverage is the fraction of all 32 units whose temporal median entry-route improvement over factual is positive.

Table B2. Frozen cell-level requirements for the complete structured-GRU transport/function assay.

| Requirement | Frozen rule |
| --- | --- |
| Contrastable coverage | >= 0.75 over 320 registered unit-times |
| Positive-unit coverage | >= 0.75 over all 32 units |
| Transported-image capture | Nested median strictly > 0.80 |
| Functional recovery | Nested contrastable median strictly > 0.50 |
| Tangent vs. factual error | Same masked nested tangent median strictly < factual median |
| Each control: median separation | Nested median of control-minus-tangent raw error difference strictly > 0 |
| Each control: correct-win fraction | Strict win fraction over all valid unit-times strictly > 0.75 |
| Control inference | All six within-checkpoint Holm-adjusted signed-rank p-values strictly < 0.05 |
| Numerical/record validity | All required records and numerical checks valid |
| Checkpoint support | Complete cell passes in both S1 and S2 |

For recovery, each unit contributes the median over only its contrastable releases, and the outer median includes only units with at least one contrastable release; an empty contrastable population is non-evaluable. Recovery is not clipped. The sign-flip, wrong-object, and in-plane controls are evaluated on all registered numerically valid unit-times without contrastability filtering. Planned inference uses one temporal-median control-minus-tangent difference per unit, a two-sided exact conditional signed-rank test that discards exact zeros and averages exact absolute-value ties, followed by Holm correction jointly across the six stratum-by-control tests in each checkpoint. No wrong-time control is part of this complete-assay conjunction.

A complete cell must satisfy every coverage, capture, recovery, raw-error, control, statistical, and numerical requirement; a checkpoint passes only if both strata pass. Invalid or missing required records are not silently excluded. The original three-checkpoint panel additionally required at least two complete checkpoints and, separately in each stratum, an across-checkpoint capture median above 0.80 and a positive median effect for every sham. The later fixed-sample extension introduced no new binary prevalence threshold. These frozen rules reproduce the reported 18/18 capture/function subtotal, 16/18 complete cells, 7/9 baseline-eligible checkpoints, and 7/10 fixed-sample checkpoints.

## B.6 Monolithic LSTM privileged rank-6 transport

The LSTM comparison tests whether the transport–function phenomenon observed in the structured-GRU model persists under a different recurrent organization. Because the structured-GRU carrier interface is architecture-specific, we did not require the LSTM comparator to reproduce the same $\mathbf{U}_4$ interface or its external address map. Instead, a development-only privileged capacity assay was used to identify a compact checkpoint-specific correction that was sufficient to initialize the transport test. The native-counterfactual anchor difference was projected into checkpoint-specific candidate carrier bases. All three development models supported tested ranks 6, 8, 12, and 16, whereas none supported ranks 1, 2, or 4. Rank 6 was therefore frozen before fresh evaluation as the smallest common tested privileged-capacity rank. This is a minimum on the tested development grid, not an estimate of intrinsic or closed-state dimension. The structured-GRU RSSM schematic is the same as in Paper 1, whereas the present appendix supplies the LSTM comparator schematic in Figure B1.

After rank 6 was frozen, a checkpoint-specific basis $\mathbf{U}_6^L \in \mathbb{R}^{384\times 6}$ was fitted for each fresh checkpoint on the fixed development fit population, without using fresh final outcomes. The privileged entry correction is

$$\boldsymbol{\delta}_0^O = \mathbf{U}_6^L(\mathbf{U}_6^L)^\top\left(\mathbf{z}_{t,0}^{CF,L} - \mathbf{z}_{t,0}^{F,L}\right) \qquad \mathbf{z}_{t,0}^{O,L} = \mathbf{z}_{t,0}^{F,L} + \boldsymbol{\delta}_0^O \tag{B15}$$

The executed tangent seed is the float32 projected vector in the first expression of Eq. (B15), evaluated before its addition to the factual carrier. It is not recomputed from the finite-precision subtraction $\mathbf{z}_{t,0}^{O,L} - \mathbf{z}_{t,0}^{F,L}$. The architecture-specific factual Jacobian acts on the complete hidden-plus-cell carrier, $\mathbf{J}_k^{F,L} \in \mathbb{R}^{384\times 384}$, and defines $\boldsymbol{\Phi}_{k\leftarrow 0}^{F,L}$ exactly as in Eq. (B3). The projected correction and its six basis columns are propagated by

$$\hat{\boldsymbol{\delta}}_k^{F,L} = \boldsymbol{\Phi}_{k\leftarrow 0}^{F,L}\boldsymbol{\delta}_0^O \qquad \mathbf{U}_{6,k}^{tr,L} = \boldsymbol{\Phi}_{k\leftarrow 0}^{F,L}\mathbf{U}_6^L \qquad \mathbf{Q}_k^L = \mathrm{orth}\left(\mathbf{U}_{6,k}^{tr,L}\right) \tag{B16}$$

No later privileged or native-counterfactual state enters this tangent chain. Geometric capture uses the LSTM-specific transported image $\mathbf{Q}_k^L$, and functional restart uses Eq. (B11) with $\hat{\boldsymbol{\delta}}_k^{F,L}$. The formal panel is fixed at fresh checkpoints 392001, 392002, and 392003. Each checkpoint uses the same 64-unit registry structure, with 32 S1 and 32 S2 units, and releases $k = 1, \dots, 10$.

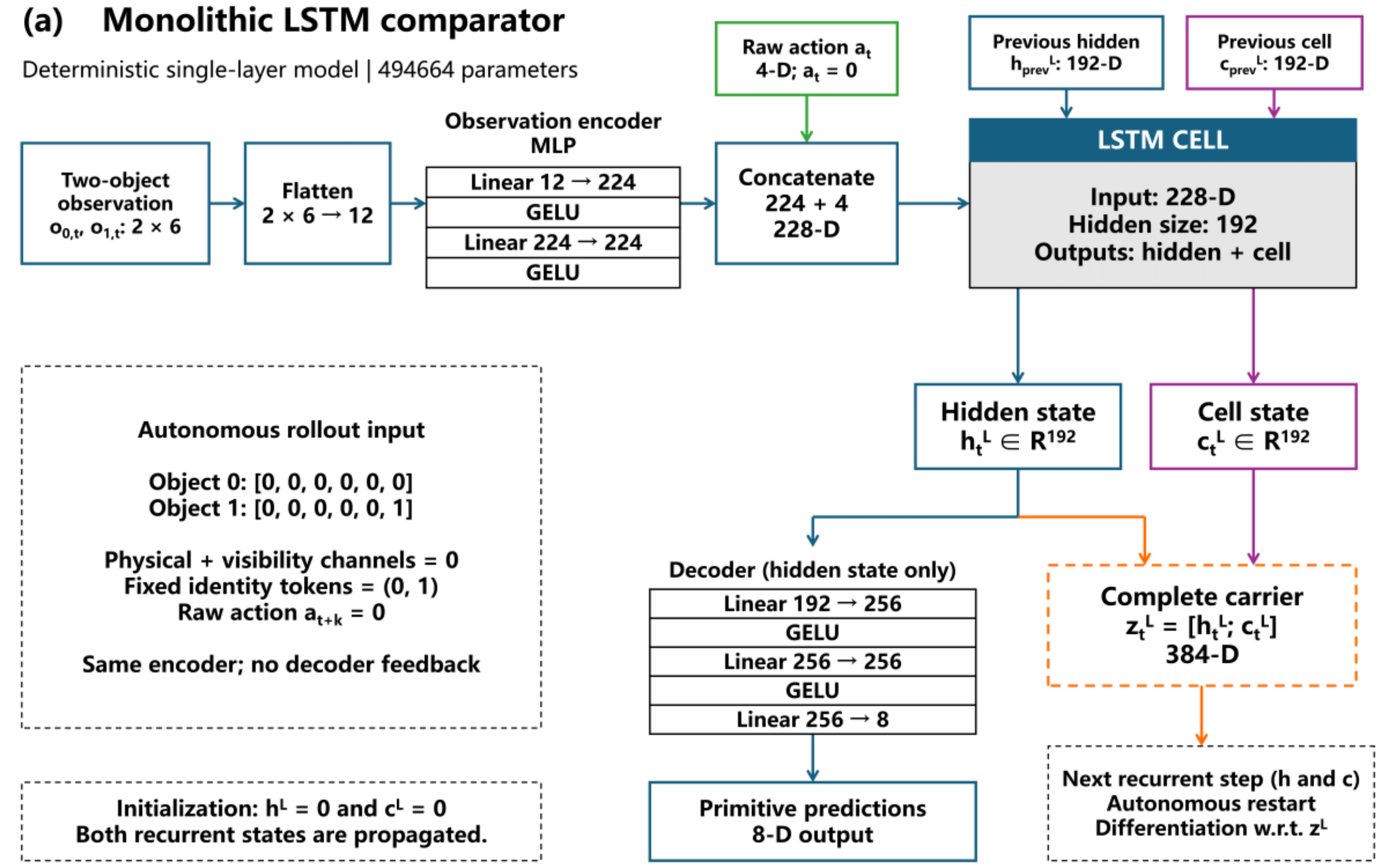

**(b) How the 228-D input updates the 192-D hidden and cell states**

Input and hidden state use separate learned projections; the cell state follows the gated memory path.

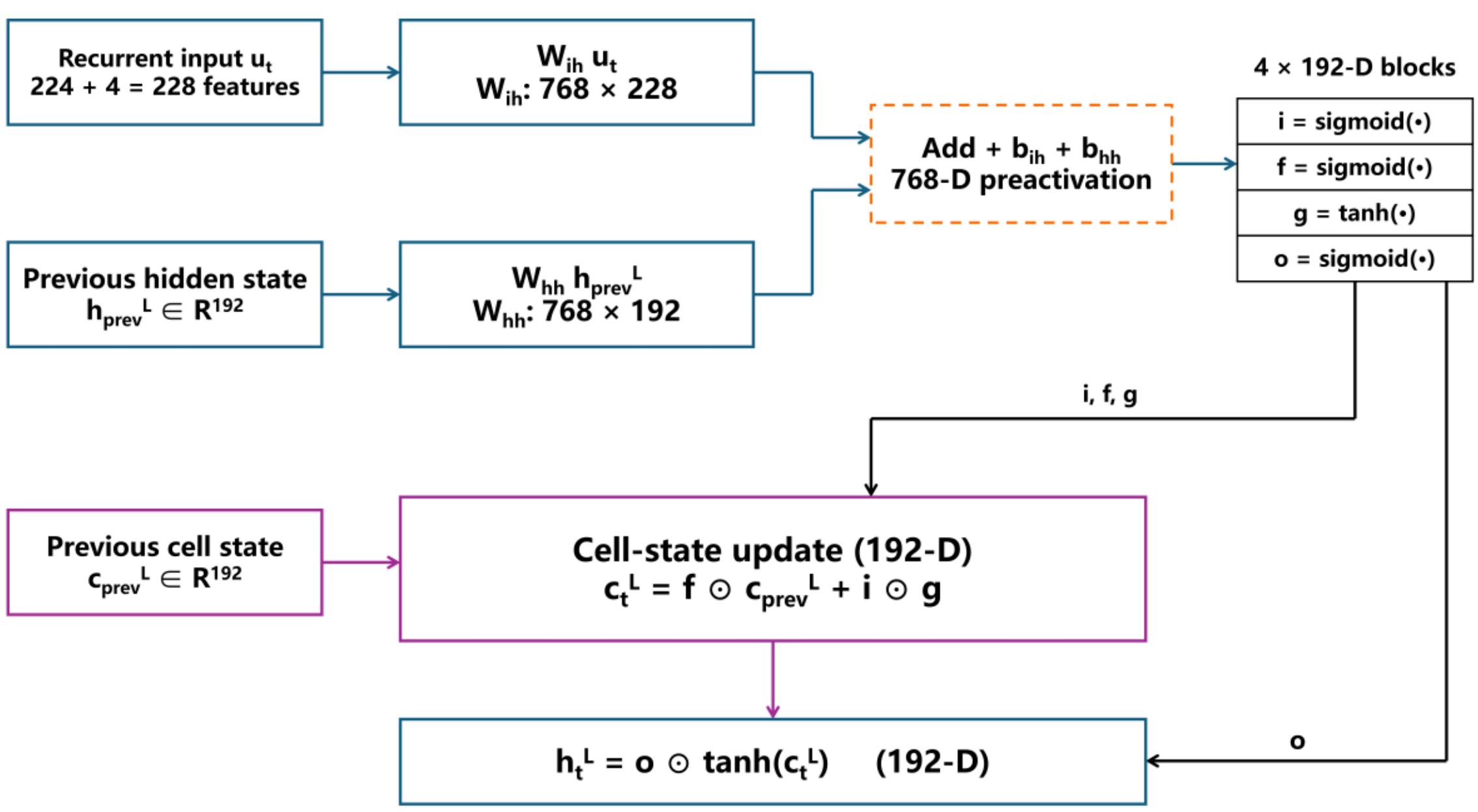

u is diagram shorthand for the concatenated input; ⊙ denotes elementwise multiplication.

Figure B1 (a) Architecture of the monolithic LSTM comparator. The two-object observation is flattened into a 12-dimensional vector and encoded by an MLP with two 224-dimensional layers and GELU activations. The resulting features are concatenated with the 4-dimensional action to form the 228-dimensional LSTM input. The recurrent cell maintains a 192-dimensional hidden state $\boldsymbol{h}_t^L$ and a 192-dimensional cell state $\boldsymbol{c}_t^L$. An MLP decoder maps $\boldsymbol{h}_t^L$ to eight primitive predictions, whereas the complete recurrent carrier $\boldsymbol{z}_t^L = [\boldsymbol{h}_t^L; \boldsymbol{c}_t^L] \in \mathbb{R}^{384}$ is retained for state propagation, autonomous restart, and differentiation. Both recurrent states are initialized to zero. During autonomous rollout, physical and visibility channels and actions are zeroed, fixed object-identity tokens are retained, and predictions are not fed back into the encoder. (b) Input projections and gated state updates within the LSTM cell. The 228-dimensional input $\boldsymbol{u}_t$ and previous 192-dimensional hidden state are separately projected by $\boldsymbol{W}_{ih} \in \mathbb{R}^{768\times228}$ and $\boldsymbol{W}_{hh} \in \mathbb{R}^{768\times192}$. Their sum, including bias terms, is partitioned into four 192-dimensional blocks that produce the input gate $\boldsymbol{i}$, forget gate $\boldsymbol{f}$, candidate update $\boldsymbol{g}$, and output gate $\boldsymbol{o}$. The cell and hidden states are updated as $\boldsymbol{c}_t^L = \boldsymbol{f} \odot \boldsymbol{c}_{prev}^L + \boldsymbol{i} \odot \boldsymbol{g}$ and $\boldsymbol{h}_t^L = \boldsymbol{o} \odot tanh(\boldsymbol{c}_t^L)$, respectively. Here, $\boldsymbol{u}_t$ denotes the concatenated observation features and action, and $\odot$ denotes elementwise multiplication.

For the LSTM assay, a release is contrastable when the privileged entry route improves remaining-future error over the factual route by more than 10^-5 (Eq. B13 with the entry route set to the privileged route). Contrastable coverage is the number of such releases divided by all 32 × 10 registered unit-times in the stratum. A positive unit is one whose temporal median factual-minus-entry error over all ten releases is positive; all 32 registered units remain in this denominator.

Transported-image capture is summarized by taking the median over all ten releases within each unit and then the median across all 32 units, without a contrastability mask. Recovery first takes the median over contrastable releases within each unit and then the median across units having at least one contrastable release. Units with no contrastable release are omitted only from this recovery median; they remain in the coverage denominators. The tangent and factual error medians use the same contrastable release/unit mask.

Each checkpoint-stratum must satisfy contrastable coverage >= 0.75, positive-unit coverage >= 0.75, nested capture > 0.80, nested contrastable recovery > 0.50, and nested tangent error < factual error, together with complete numerical, registry, and autonomy validity. Numerical validity requires finite errors, states, and decodes, strict carrier/decoded norm and component bounds below $10^6$, complete registered release coverage, unchanged weights, and lossless artifact readback. Both S1 and S2 are required for checkpoint support. The LSTM panel has no primary semantic mapper or semantic sham gate, and no rank rescue, mapper retuning, or seed replacement is allowed. All three fresh checkpoints satisfy this conjunction.

## Appendix C. Finite-horizon full-carrier future-response operator

### C.1 Autonomous future map

For each matched route $q \in \{F, P, CF\}$, let the route-specific carrier at the intervention anchor be $\mathbf{z}_t^{(q)}$. Starting from this carrier, the released model evolves according to

$$\mathbf{z}_{t,0}^{(q)} = \mathbf{z}_t^{(q)} \qquad \mathbf{z}_{t,k+1}^{(q)} = T\left(\mathbf{z}_{t,k}^{(q)}, \mathbf{a}_{t+k}\right) \qquad k = 0, \dots, H-1 \tag{C1}$$

Here, $q$ labels the factual, patched, or native-counterfactual route; $\mathbf{z}_{t,k}^{(q)} \in \mathbb{R}^{192}$ is the complete carrier obtained after $k$ autonomous transitions from the anchor of route $q$; $T$ is one frozen recurrent transition of the trained world model without observation assimilation; $\mathbf{a}_{t+k}$ is the recorded action supplied to transition $k$; and $H$ is the future-response horizon. The assay uses $H = 8$, and all corresponding action entries are zero.

Let $D: \mathbb{R}^{192} \to \mathbb{R}^8$ denote the decoder from the complete carrier to the two-object primitive coordinates, and let $\mathbf{w} \in \mathbb{R}^8$ be the frozen primitive-coordinate scaling vector used by the response assay. The weighted future decode is

$$\mathbf{y}_{t,k}^{(q)} = \mathbf{w} \odot D\left(\mathbf{z}_{t,k}^{(q)}\right) \qquad k = 1, \dots, H \tag{C2}$$

Here, $\mathbf{y}_{t,k}^{(q)} \in \mathbb{R}^8$ is the weighted primitive-state decode after $k$ released transitions; $\mathbf{w}$ is the frozen eight-coordinate scaling vector inherited from the Paper 1 M3 output metric, with each entry given by the corresponding decoder denormalization standard deviation divided by the frozen primitive-coordinate scale. The same $\mathbf{w}$ is applied at every future step so that position and velocity coordinates contribute on the same predefined scale. $D$ is the frozen decoder; and $\odot$ denotes elementwise multiplication.

The stacked future map is then

$$G_{t,H}\left(\mathbf{z}_t^{(q)}\right) = \mathrm{vec}_{\mathrm{step}}\left(\mathbf{y}_{t,1}^{(q)}, \dots, \mathbf{y}_{t,H}^{(q)}\right) = \mathbf{Y}_{t,H}^{(q)} \in \mathbb{R}^{8H} \tag{C3}$$

Here, $\mathrm{vec}_{\mathrm{step}}$ denotes concatenation of the weighted future decodes in future-step-major order, and $\mathbf{Y}_{t,H}^{(q)}$ is the resulting weighted future-output vector. The anchor decode is not included. With $H = 8$, the vector has dimension $8H = 64$, so $\mathbf{Y}_{t,H}^{(q)} \in \mathbb{R}^{64}$.

The future map uses the eight action slots beginning at the anchor, corresponding to the frozen action slice 7:15 in the main $t = 7$ assay. No future observation is read during this map, and there is no teacher forcing, decoder feedback, repeated hidden intervention, or carrier clamping.

### C.2 Full-carrier response operator and local approximation

The route-specific future-response operator is the Jacobian of the complete future map with respect to its 192-dimensional carrier input:

$$\mathbf{R}_{t,H}^{(q)} = \left.\frac{\partial G_{t,H}(\mathbf{z})}{\partial \mathbf{z}}\right|_{\mathbf{z}=\mathbf{z}_t^{(q)}} \in \mathbb{R}^{8H\times 192}$$

(C4)

Here, $G_{t,H}$ is the frozen future map defined in Eq. (C3), and $\mathbf{R}_{t,H}^{(q)}$ is the resulting Jacobian evaluated at the factual, patched, or native-counterfactual anchor carrier. For $H = 8$, its shape is $64 \times 192$. The derivative is taken with respect to the complete carrier rather than the four-dimensional intervention coefficients or the rank-4 carrier subspace.

For a local perturbation $\delta\mathbf{z}_t$,

$$G_{t,H}\left(\mathbf{z}_t^{(q)} + \delta\mathbf{z}_t\right) - G_{t,H}\left(\mathbf{z}_t^{(q)}\right) = \mathbf{R}_{t,H}^{(q)}\,\delta\mathbf{z}_t + o(\parallel \delta\mathbf{z}_t \parallel_2)$$

(C5)

Here, $\delta\mathbf{z}_t \in \mathbb{R}^{192}$ is an infinitesimal perturbation of the route-specific complete carrier; the left-hand side is the corresponding finite difference in the stacked weighted future output; and the little-$o$ term is negligible relative to $\parallel \delta\mathbf{z}_t \parallel_2$ as $\parallel \delta\mathbf{z}_t \parallel_2 \to 0$. Equation (C5) is a local first-order relation; it does not assert that the same linear approximation remains accurate at the finite amplitude of the inherited Paper 1 patch.

### C.3 Chain-rule interpretation

The full future-response operator can be interpreted as the composition of local recurrent propagation and output sensitivity along the released trajectory.

Define the one-step recurrent Jacobian by

$$\mathbf{J}_{t,k}^{(q)} = \left.\frac{\partial T(\mathbf{z}, \mathbf{a}_{t+k})}{\partial \mathbf{z}}\right|_{\mathbf{z}=\mathbf{z}_{t,k}^{(q)}} \in \mathbb{R}^{192\times 192} \qquad k = 0, \dots, H-1$$

(C6)

Here, $\mathbf{J}_{t,k}^{(q)}$ is the Jacobian of one autonomous recurrent transition with respect to the complete carrier; $k$ indexes the released transition; $T$ is the frozen transition map; $\mathbf{a}_{t+k}$ is the fixed action at that transition; and $\mathbf{z}_{t,k}^{(q)}$ is the carrier along route $q$ at which the derivative is evaluated.

Define the weighted-decoder Jacobian by

$$\mathbf{C}_{t,k}^{(q)} = \left.\frac{\partial[\mathbf{w} \odot D(\mathbf{z})]}{\partial \mathbf{z}}\right|_{\mathbf{z}=\mathbf{z}_{t,k}^{(q)}} \in \mathbb{R}^{8\times 192} \qquad k = 1, \dots, H$$

(C7)

Here, $\mathbf{C}_{t,k}^{(q)}$ describes how a perturbation of the complete carrier after $k$ released transitions changes the weighted eight-dimensional decoder output at that step; $\mathbf{w}$ is the frozen output-weight vector; and $D$ is the frozen decoder.

The finite-step carrier propagator is

$$\mathbf{\Phi}_{t,0}^{(q)} = \mathbf{I}_{192} \qquad \mathbf{\Phi}_{t,k}^{(q)} = \mathbf{J}_{t,k-1}^{(q)}\mathbf{J}_{t,k-2}^{(q)} \cdots \mathbf{J}_{t,0}^{(q)} \qquad k = 1, \dots, H$$

(C8)

Here, $\mathbf{\Phi}_{t,k}^{(q)} \in \mathbb{R}^{192\times192}$ maps an infinitesimal perturbation of the anchor carrier to its first-order effect on the carrier after $k$ autonomous transitions, and $\mathbf{I}_{192}$ is the $192 \times 192$ identity matrix. The multiplication order follows the temporal order of recurrent transitions from the anchor to future step $k$.

By the chain rule,

$$\mathbf{R}_{t,H}^{(q)} = \begin{bmatrix} \mathbf{C}_{t,1}^{(q)}\mathbf{\Phi}_{t,1}^{(q)} \\ \mathbf{C}_{t,2}^{(q)}\mathbf{\Phi}_{t,2}^{(q)} \\ \vdots \\ \mathbf{C}_{t,H}^{(q)}\mathbf{\Phi}_{t,H}^{(q)} \end{bmatrix} \tag{C9}$$

Here, each block row $\mathbf{C}_{t,k}^{(q)}\mathbf{\Phi}_{t,k}^{(q)} \in \mathbb{R}^{8\times192}$ gives the first-order dependence of the weighted decoder output at future step $k$ on the anchor carrier. Stacking the $H$ block rows yields the full $8H \times 192$ future-response operator.

Equation (C9) makes explicit that the response operator combines repeated recurrent propagation with the sensitivity of each released decoded state to the carrier. This decomposition is a mathematical interpretation of the derivative. The production quantity used in the analysis is obtained by differentiating the complete frozen future map in Eq. (C3) with respect to its carrier input; separate estimation or fitting of the matrices in Eq. (C9) is not required.

### C.4 Route matching and numerical boundary

For every evaluation unit and checkpoint, $\mathbf{R}_{t,H}^{(F)}$, $\mathbf{R}_{t,H}^{(P)}$, and $\mathbf{R}_{t,H}^{(CF)}$ are evaluated at their respective post-assimilation or post-patch anchor carriers. Eq. (C4) differentiates only the future response from a fixed anchor carrier; the construction of that anchor carrier is not part of the derivative.

All three route-specific operators use the same trained checkpoint, future action sequence, horizon, decoder, primitive-coordinate weighting, output ordering, and autonomous-release convention. No projection onto $\mathbf{U}_4$ or any other low-dimensional subspace is applied before differentiation. Thus, each operator describes sensitivity with respect to all 192 coordinates of the assembled carrier.

The frozen $H = 8$ assay gives a $64 \times 192$ operator because it queries eight primitive coordinates at each of eight future states. The 64-dimensional output side is determined by the selected future-output query and must not be interpreted as a 64-dimensional model state.

The response operator is not fitted from trajectory pairs. It is a local derivative of the frozen recurrent model at a specified route-specific carrier. Model parameters remain fixed throughout the calculation.

### C.5 Interpretation boundary

$\mathbf{R}_{t,H}^{(q)}$ is a trajectory-conditioned finite-horizon sensitivity operator. It identifies which infinitesimal perturbations of the current full carrier can influence the selected weighted future-output query, and how those effects propagate over the eight released transitions. Its singular geometry therefore describes local future sensitivity rather than named physical coordinates. We assess the functional relevance of this local structure through matched route comparisons, causal perturbation tests, and finite-dose validation.

## Appendix D. Future-sensitive subspaces and alignment metrics

This appendix defines the singular-spectrum, right-subspace, and full-operator quantities used to compare the matched factual, patched, and native-counterfactual future-response operators from Appendix C.

### D.1 Native-counterfactual sensitivity rank

For route $q \in \{\mathrm{F}, \mathrm{P}, \mathrm{CF}\}$, write the compact singular-value decomposition of the response operator as

$$\mathbf{R}_{t,H}^{(q)} = \mathbf{L}^{(q)}\mathbf{\Sigma}^{(q)}\left(\mathbf{V}^{(q)}\right)^{\mathsf{T}}$$

(D1)

Here, $\mathbf{R}_{t,H}^{(q)} \in \mathbb{R}^{64\times 192}$ is the route-specific future-response operator; $\mathbf{L}^{(q)}$ contains its left singular vectors; $\mathbf{\Sigma}^{(q)}$ contains its nonnegative singular values in descending order; $\mathbf{V}^{(q)}$ contains the corresponding right singular vectors in the 192-dimensional carrier space; and $\mathsf{T}$ denotes matrix transpose. Because the output dimension is 64, the compact decomposition contains at most 64 nonzero singular values.

Let $\sigma_i^{(\mathrm{CF})}$ be the $i$-th singular value of the native-counterfactual operator, and let $m \le 64$ denote the number of singular values represented in its compact decomposition. The matched comparison rank is

$$r_{\mathrm{CF}} = \min\left\{k: \frac{\sum_{i=1}^{k}\left(\sigma_i^{(\mathrm{CF})}\right)^2}{\sum_{i=1}^{m}\left(\sigma_i^{(\mathrm{CF})}\right)^2} \ge 0.90\right\}$$

(D2)

Here, $r_{\mathrm{CF}}$ is the smallest native-counterfactual rank reaching $90\%$ of the squared singular-value sum; $k$ is a candidate truncation rank; $i$ indexes singular values; and $m$ is the compact-spectrum length. The denominator is required to be positive. The same unit-specific $r_{\mathrm{CF}}$ is reused for all matched factual-native-CF, patched-native-CF, and patched-factual subspace comparisons. Ranks obtained separately from the factual or patched spectra are descriptive only and do not set the comparison dimension. In particular, $r_{\mathrm{CF}}$ is not an intrinsic-state, recurrent-state, or effective-state dimension.

### D.2 Right-singular subspaces and principal-angle similarity

For route $A \in \{\mathrm{F}, \mathrm{P}, \mathrm{CF}\}$, let $\mathbf{V}_{A,r} = \left[\mathbf{v}_{A,1} \cdots \mathbf{v}_{A,r}\right] \in \mathbb{R}^{192\times r}$ denote the column-orthonormal basis formed from the first $r$ right singular vectors of $\mathbf{R}_{t,H}^{(A)}$. For two matched routes $A$ and $B$, let $\kappa_j^{A,B;r}$ be the $j$-th singular value of $\mathbf{V}_{A,r}^{\mathsf{T}}\mathbf{V}_{B,r}$. Numerically, these singular values are clipped to $[0,1]$ before squaring. The subspace similarity is

$$S_{\mathrm{sub}}^{A,B;r} = \frac{1}{r}\sum_{j=1}^{r}\mathrm{clip}\left(\kappa_j^{A,B;r}, 0,1\right)^2 = \frac{1}{r}\sum_{j=1}^{r}\cos^2\left(\theta_j^{A,B}\right)$$

(D3)

Here, $\mathbf{v}_{A,j} \in \mathbb{R}^{192}$ is the $j$-th right singular vector for route $A$; $r$ is the common subspace dimension; $\kappa_j^{A,B;r}$ is the numerically stabilized cosine of the $j$-th principal angle; $\theta_j^{A,B}$ is that principal angle; and $\mathrm{clip}(x, 0,1)$ truncates numerical values below zero to zero and above one to one. The production implementation stores the transpose-equivalent orthonormal row-basis representation; Eq. (D3) uses column-basis notation for clarity.

The manuscript comparisons use $r = r_{\mathrm{CF}}$ from Eq. (D2). Thus, $S_{\mathrm{sub}}$ compares the orientation of matched future-sensitive carrier subspaces under one common native-CF-derived dimension. It does not compare singular-value magnitudes or the output-side singular vectors.

### D.3 Patched-factual separation and CF-directed alignment

Once $S_{\mathrm{sub}}$ is defined, we use two contrasts that answer different questions. The direct patched-factual separation is

$$\Delta_{\text{PF-Align}} = 1 - S_{\mathrm{sub}}^{\mathrm{P,F};r_{\mathrm{CF}}} \tag{D4}$$

Here, $\Delta_{\text{PF-Align}}$ is the separation between the patched and factual leading future-sensitive subspaces under the same native-CF-derived comparison rank. A value of zero means that the tested patched and factual subspaces coincide under this overlap measure; larger values indicate less overlap.

The formal primary alignment contrast is

$$\Delta_{\mathrm{Align}} = S_{\mathrm{sub}}^{\mathrm{P,CF};r_{\mathrm{CF}}} - S_{\mathrm{sub}}^{\mathrm{F,CF};r_{\mathrm{CF}}} \tag{D5}$$

Here, $\Delta_{\mathrm{Align}}$ is the unit-level CF-directed alignment gain. Positive values mean that, under the same native-CF-derived comparison rank, the patched leading future-sensitive subspace is more aligned with native CF than the factual one is. Unlike $\Delta_{\text{PF-Align}}$, this quantity asks about orientation relative to native CF rather than the magnitude of patched-factual subspace separation.

The two quantities do not form a Euclidean decomposition, and the ratio $\Delta_{\mathrm{Align}}/\Delta_{\text{PF-Align}}$ must not be interpreted as the fraction of patched-factual change directed toward native CF. $\Delta_{\text{PF-Align}}$ is descriptive and non-gating; formal alignment decisions use $\Delta_{\mathrm{Align}}$ under the checkpoint-first rule specified in the statistical protocol. Checkpoint–stratum medians of both contrasts are reported in Table H2 (Appendix H.3).

### D.4 Full-operator change and CF-directed operator comparison

To distinguish leading input-side subspace orientation from changes in the complete input-to-future-output mapping, we use the symmetric normalized Frobenius discrepancy

$$d_R^{A,B} = \frac{\left\| \mathbf{R}_{t,H}^{(A)} - \mathbf{R}_{t,H}^{(B)} \right\|_F}{\sqrt{\left\| \mathbf{R}_{t,H}^{(A)} \right\|_F \left\| \mathbf{R}_{t,H}^{(B)} \right\|_F + 10^{-12}}} \tag{D6}$$

Here, $d_R^{A,B}$ is the normalized distance between the complete response operators for routes $A$ and $B$; $\|\cdot\|_F$ denotes the Frobenius norm; and $10^{-12}$ is the frozen numerical floor in the denominator. In particular, $d_R^{\mathrm{P,F}}$ provides a descriptive measure of the amount of patched-factual full-operator change. Unlike the subspace quantities, $d_R^{A,B}$ is sensitive to the complete response mapping, including singular magnitudes, output-side geometry, and non-leading input directions.

The supporting CF-directed full-operator contrast is

$$\Delta_d = d_R^{\mathrm{F,CF}} - d_R^{\mathrm{P,CF}}$$

(D7)

Here, $\Delta_d$ is positive when the patched full response operator is closer to the native-counterfactual operator than the factual operator is under the registered $d_R$ normalization. The roles of $d_R^{\mathrm{P,F}}$ and $\Delta_d$ parallel those of $\Delta_{\text{PF-Align}}$ and $\Delta_{\text{Align}}$, respectively, but they operate on the complete response matrices. $\Delta_d$ is supporting only and cannot replace or rescue a failed primary $\Delta_{\text{Align}}$ criterion.

### D.5 Metric status and interpretation boundary

The quantities above provide a two-level description of matched future-response change. $\Delta_{\text{Align}}$ is the primary alignment metric; $\Delta_d$ is supporting; and $\Delta_{\text{PF-Align}}$ and $d_R^{\mathrm{P,F}}$ are descriptive. Their different roles are fixed before checkpoint-level interpretation, and supporting or descriptive quantities do not replace the primary criterion.

All comparisons remain specific to the tested checkpoint, matched unit, dynamical stratum, eight-transition output query, primitive-coordinate weighting, and native carrier metric. Similarity of the tested leading subspaces does not imply identical internal computation. None of these quantities establishes a low-dimensional recurrent state, dynamical closure, a minimal realization, or an effective-state dimension.

## Appendix E. Local future-response sensitivity and perturbation validation

This appendix defines the frozen direction-construction rules and route-specific panels, matched perturbation scale, finite autonomous releases, and unit-level rank-agreement statistic used to validate the local future-response operator from Appendix C.

### E.1 Frozen causal-validation subset and direction bank

The perturbation validation uses a frozen 32-unit subset at each checkpoint, with 16 units from each registered dynamical stratum. Each unit is evaluated separately at the patched and native-counterfactual anchor carriers. At each basepoint, 52 unit-norm directions are generated using the same frozen direction families, ordering, and construction rules. For a fixed checkpoint and matched unit, the route-specific panel is

$$\Xi^{(q)} = \left\{\boldsymbol{\xi}_j^{(q)}\right\}_{j=1}^{52} \qquad q \in \{\mathrm{P}, \mathrm{CF}\} \qquad \boldsymbol{\xi}_j^{(q)} \in \mathbb{R}^{192} \qquad \left\|\boldsymbol{\xi}_j^{(q)}\right\|_2 = 1 \tag{E1}$$

Here, $\Xi^{(q)}$ is the set of 52 perturbation directions constructed for a given checkpoint, matched unit, and route. Each $\xi_j^{(q)}$ is a 192-dimensional vector with unit Euclidean norm, specifying a direction of change in the full carrier space. The label $q \in \{P, CF\}$ denotes the patched or native-counterfactual basepoint, and $j = 1, \ldots, 52$ indexes the directions. Both routes follow the same ordering of the six direction families described below.

Table E1. Composition of the 52-direction perturbation panel for one route and matched unit. The first four families are route-specific operator-derived directions (4 each), the Paper 1 interface contributes four shared directions, and the projected-random family contributes 32 route-specific directions.

| Direction family | Count | Construction |
|---|---|---|
| Leading future response | 4 | First four right singular vectors of $\mathbf{R}_{t,H}^{(q)}$. |
| Null future response | 4 | Last four vectors of the full 192-dimensional right SVD basis of $\mathbf{R}_{t,H}^{(q)}$. |
| Leading carrier propagation | 4 | First four right singular vectors of the eight-step propagator $\boldsymbol{\Phi}_{t,H}^{(q)}$. |
| Leading anchor decoder | 4 | First four right singular vectors of the anchor decoder Jacobian. |
| Paper 1 interface | 4 | Individually normalized columns of the checkpoint-specific $\mathbf{U}_4$. |
| Projected random | 32 | Deterministic Gaussian draws projected off the leading $r_{\mathrm{CF}}$-dimensional right-singular subspace of $\mathbf{R}_{t,H}^{(q)}$, then normalized. |

The anchor-decoder family uses the derivative of the decoder $D$ at the route-specific anchor without the future-output scaling $\mathbf{w}$; it is distinct from the weighted future decoder Jacobian in Appendix C. The null family uses the full right SVD basis of the 64-by-192 response operator, rather than the smallest nonzero singular modes.

For the random family, $\mathbf{V}_{q,r_{\mathrm{CF}}}$ contains the leading $r_{\mathrm{CF}}$ right singular vectors of the route-specific response operator. The rank $r_{\mathrm{CF}}$ is defined from the native-CF spectrum in Appendix D and is shared by the matched routes. Each standard-normal draw $\boldsymbol{\eta} \in \mathbb{R}^{192}$ is projected as follows and then normalized to unit length:

$$\widetilde{\boldsymbol{\xi}_j^{(q)}} = \left(\mathbf{I} - \mathbf{V}_{q,r_{\mathrm{CF}}} {\mathbf{V}_{q,r_{\mathrm{CF}}}}^{\mathrm{T}}\right)\boldsymbol{\eta} \tag{E2}$$

Equation (E2) defines the unnormalized projected draw. The Gaussian draw is 192-dimensional, and the projection removes its component inside the leading future-response subspace at the selected route, using the common native-CF-derived comparison rank. The implementation then normalizes the projected vector to unit Euclidean norm before storing it as the registered direction.

The 32 projected-random directions are generated independently and are not mutually orthogonalized.

**Deterministic seed construction.** For projected-random direction indices 0–31, NumPy default_rng is seeded with the unsigned little-endian integer represented by the first eight bytes of the SHA-256 digest of the UTF-8 string

```
P2_FFD_FRESH_CAUSAL_V1|{checkpoint}|{unit_id}|{route}|{direction_index}
```

The route token is part of the seed, so the projected-random vectors are route-specific even for the same checkpoint, matched unit, and direction index.

**Panel sharing across routes.** The 16 operator-derived directions and 32 projected-random directions are constructed separately at the patched and native-counterfactual basepoints. The four normalized Paper 1 interface directions are shared. Thus the two routes use the same six-family composition and index ordering, but not 52 identical coordinate vectors. All direction vectors are fixed by the registered construction before their finite rollout effects are measured.

### E.2 Matched perturbation scale and autonomous releases

For each matched unit, the finite perturbation amplitude is tied to the norm of the inherited full patched-minus-factual anchor correction,

$$\Delta\mathbf{z}_t^{(P)} = \mathbf{z}_t^{(P)} - \mathbf{z}_t^{(F)} \qquad \epsilon = 0.10\,\left\|\Delta\mathbf{z}_t^{(P)}\right\|_2 \tag{E3}$$

Here, $\Delta\mathbf{z}_t^{(P)} \in \mathbb{R}^{192}$ is the actual one-shot patched-minus-factual carrier change, and $\epsilon$ is the positive scalar perturbation magnitude shared by all registered directions of the matched unit. The same $\epsilon$ is used for both perturbation signs and for the patched and native-counterfactual basepoints; the vectors $\boldsymbol{\xi}_j^{(q)}$ follow the route-specific construction in E.1. This 10% scale is distinct from the full inherited patch analyzed in Section 3.6.

For a base route $q \in \{\mathrm{P}, \mathrm{CF}\}$, the model is released from $\mathbf{z}_t^{(q)}$, $\mathbf{z}_t^{(q)} + \epsilon\boldsymbol{\xi}_j^{(q)}$, and $\mathbf{z}_t^{(q)} - \epsilon\boldsymbol{\xi}_j^{(q)}$ under the same frozen future action sequence. Every release uses the same $H = 8$ future-output query as Appendix C. No future observations, teacher forcing, decoder feedback, repeated intervention, physical-state clamping, or hidden-state clamping are introduced during these releases.

### E.3 Predicted and measured direction effects

The operator-predicted effect for registered direction $\boldsymbol{\xi}_j^{(q)}$ at route $q$ is

$$g_{\mathrm{pred}}\left(\boldsymbol{\xi}_j^{(q)}; q\right) = \left\|\mathbf{R}_{t,H}^{(q)}\boldsymbol{\xi}_j^{(q)}\right\|_2 \tag{E4}$$

Here, $\mathbf{R}_{t,H}^{(q)} \in \mathbb{R}^{64\times 192}$ is the route-specific future-response operator and $\mathbf{R}_{t,H}^{(q)}\boldsymbol{\xi}_j^{(q)} \in \mathbb{R}^{64}$ is its directional derivative in the weighted future-output coordinates. For perturbation amplitude $\epsilon$, the corresponding first-order change is $\epsilon$ times this vector.

The measured finite autonomous effect on the same route-specific vector is

$$g_{\text{actual}}\left(\boldsymbol{\xi}_j^{(q)};q\right)=\frac{\left\|G_{t,H}\left(\mathbf{z}_t^{(q)}+\epsilon\boldsymbol{\xi}_j^{(q)}\right)-G_{t,H}\left(\mathbf{z}_t^{(q)}\right)\right\|_2+\left\|G_{t,H}\left(\mathbf{z}_t^{(q)}-\epsilon\boldsymbol{\xi}_j^{(q)}\right)-G_{t,H}\left(\mathbf{z}_t^{(q)}\right)\right\|_2}{2\epsilon}$$

(E5)

Here, $G_{t,H}\colon \mathbb{R}^{192}\to\mathbb{R}^{64}$ is the frozen future map from Appendix C. The numerator contains the two separate future-deviation norms produced by positive and negative perturbations of the same route-specific registered direction. The implementation adds these norms and then divides by $2\epsilon$; it does not first subtract the two perturbed future vectors and take a single norm. Prediction and measurement use identical coordinates within each route, although most direction coordinates differ between routes.

### E.4 Rank-agreement statistic and interpretation boundary

For base route $q$, collect the 52 direction effects as $\mathbf{g}_{\text{pred}}^{(q)}=\left[g_{\text{pred}}\left(\boldsymbol{\xi}_j^{(q)};q\right)\right]_{j=1}^{52}$ and $\mathbf{g}_{\text{actual}}^{(q)}=\left[g_{\text{actual}}\left(\boldsymbol{\xi}_j^{(q)};q\right)\right]_{j=1}^{52}$. Their unit-level rank agreement is

$$\rho^{(q)}=\operatorname{Corr}\left(\operatorname{rank}_{\text{avg}}\left(\mathbf{g}_{\text{pred}}^{(q)}\right),\operatorname{rank}_{\text{avg}}\left(\mathbf{g}_{\text{actual}}^{(q)}\right)\right)$$

(E6)

Here, $\rho^{(q)}$ is the Spearman rank correlation; $\operatorname{rank}_{\text{avg}}$ assigns average ranks to tied values; and $\operatorname{Corr}$ correlates the two rank vectors. A constant predicted or measured rank vector is non-evaluable. The comparison is performed separately at the patched and native-counterfactual basepoints before any checkpoint- or stratum-level aggregation.

The eight-step carrier propagator $\boldsymbol{\Phi}_{t,H}^{(q)}$ from Appendix C is retained only as an auxiliary direction-panel comparator. It is not the primary perturbation-predictiveness quantity and is not used to define closure, a minimal realization, or an effective-state dimension.

The statistic compares local first-order sensitivities with separately measured finite autonomous effects on the same route-specific vectors. It is a pooled ordering test over the registered structured and operator-conditioned random panel. Because the panel is partly constructed from local operators, the result does not establish the same accuracy for arbitrary unseen directions, for each direction family separately, or for effect magnitudes. The checkpoint-first rule in Section 3.7 and Appendix G tests the patched and native-counterfactual correlations separately; it does not compare effects on identical coordinate vectors across routes.

## Appendix F. Finite-dose response reconfiguration and path accounting

This appendix defines the factual-to-patched hidden-dose path, the full-dose factual-tangent residual, the midpoint operator-chord deviation, and the numerical path-accounting checks used in the finite-dose assay.

### F.1 Hidden-patch dose path and endpoint future effect

The local perturbation assay in Appendix E uses a 10% matched-unit scale around the patched and native-counterfactual basepoints. The finite-dose assay instead follows the full inherited factual-to-patched carrier displacement. For each matched evaluation unit, define

$$\Delta\mathbf{z}_t^{(P)} = \mathbf{z}_t^{(P)} - \mathbf{z}_t^{(F)} \qquad \mathbf{z}_\gamma = \mathbf{z}_t^{(F)} + \gamma\Delta\mathbf{z}_t^{(P)} \qquad 0 \le \gamma \le 1$$

(F1)

Here, $\Delta\mathbf{z}_t^{(P)} \in \mathbb{R}^{192}$ is the full one-shot patched-minus-factual carrier displacement and $\gamma$ is a scalar hidden-patch dose. The endpoints are $\mathbf{z}_0 = \mathbf{z}_t^{(F)}$ and $\mathbf{z}_1 = \mathbf{z}_t^{(P)}$. This path is defined entirely between the factual and patched carriers. It is not a physical-edit dose path and does not interpolate toward the native-counterfactual carrier.

Using the frozen future-output map from Appendix C, define the endpoint future change

$$\Delta\mathbf{Y}_{t,H}^{(PF)} = G_{t,H}\left(\mathbf{z}_t^{(P)}\right) - G_{t,H}\left(\mathbf{z}_t^{(F)}\right) \in \mathbb{R}^{64}$$

(F2)

Here, $\Delta\mathbf{Y}_{t,H}^{(PF)}$ is the actual change in the same weighted eight-transition future-output query used throughout the response assay.

### F.2 Full-dose factual-tangent residual

Let $\mathbf{R}_\gamma = \mathbf{R}_{t,H}(\mathbf{z}_\gamma)$ denote the full-carrier future-response operator evaluated along the path. In particular, $\mathbf{R}_0 = \mathbf{R}_{t,H}^{(F)}$ and $\mathbf{R}_1 = \mathbf{R}_{t,H}^{(P)}$. A single local linearization at the factual endpoint predicts the full-dose future change as $\mathbf{R}_0\Delta\mathbf{z}_t^{(P)}$. The registered residual is

$$E_{\text{local}} = \frac{\left\|\Delta\mathbf{Y}_{t,H}^{(PF)} - \mathbf{R}_0\Delta\mathbf{z}_t^{(P)}\right\|_2}{\left\|\Delta\mathbf{Y}_{t,H}^{(PF)}\right\|_2 + 10^{-12}}$$

(F3)

Here, $E_{\text{local}}$ is dimensionless. A value of zero would mean that the factual-endpoint tangent exactly accounts for the full finite patched future under the frozen query. A nonzero value measures the residual of that one fixed local approximation; it does not by itself identify a specific nonlinearity or recurrent mechanism.

### F.3 Midpoint operator chord deviation

The pathwise response operator is

$$\mathbf{R}_\gamma = \mathbf{R}_{t,H}\left(\mathbf{z}_t^{(F)} + \gamma\Delta\mathbf{z}_t^{(P)}\right) \in \mathbb{R}^{64\times192}$$

(F4)

The registered operator-reconfiguration quantity uses only the midpoint:

$$C_R(0.5) = \frac{\|\mathbf{R}_{0.5} - 1/2\,(\mathbf{R}_0 + \mathbf{R}_1)\|_F}{\|\mathbf{R}_1 - \mathbf{R}_0\|_F + 10^{-12}}$$

(F5)

Here, $C_R(0.5)$ compares the actual midpoint operator with the straight chord between the factual and patched endpoint operators under the Frobenius geometry. It is a normalized operator-space chord deviation, not an intrinsic curvature of the recurrent-state manifold. The formal finite-dose assay does not introduce an all-$\gamma$ curvature statistic or an additional gate at other dose values.

### F.4 Path-integrated response and numerical integration

Along the fixed path in Eq. (F1), define the 64-dimensional directional future-response vector

$$\mathbf{q}_\gamma = \mathbf{R}_\gamma \Delta\mathbf{z}_t^{(P)} \in \mathbb{R}^{64}$$

(F6)

By the chain rule, $\mathbf{q}_\gamma = dG_{t,H}(\mathbf{z}_\gamma)/d\gamma$. Therefore the endpoint future change satisfies

$$\Delta\mathbf{Y}_{t,H}^{(PF)} = \int_0^1 \mathbf{q}_\gamma \, d\gamma = \int_0^1 \mathbf{R}_\gamma \, \Delta\mathbf{z}_t^{(P)} \, d\gamma$$

(F7)

This identity is used as path accounting. The frozen implementation evaluates $\mathbf{q}_\gamma$ on the 33-node grid $\gamma_j = j/32, j = 0, \dots, 32$. Nested composite-Simpson approximations are then formed for $n \in \{8,16,32\}$ subintervals using the corresponding nodes from that shared grid:

$$\mathbf{I}_n = \frac{1}{3n}\left[\mathbf{q}_0 + \mathbf{q}_1 + 4\sum_{\substack{1\le j<n \\ j \text{ odd}}} \mathbf{q}_{j/n} + 2\sum_{\substack{2\le j<n \\ j \text{ even}}} \mathbf{q}_{j/n}\right] \qquad n \in \{8,16,32\}$$

(F8)

Here, $\mathbf{I}_n \in \mathbb{R}^{64}$ is the composite-Simpson approximation to the integrated path response. The relative endpoint-accounting error is

$$E_{\text{int},n} = \frac{\left\|\Delta\mathbf{Y}_{t,H}^{(PF)} - \mathbf{I}_n\right\|_2}{\left\|\Delta\mathbf{Y}_{t,H}^{(PF)}\right\|_2 + 10^{-12}} \qquad n \in \{8,16,32\}$$

(F9)

The nested-grid convergence readbacks are

$$\text{conv}_{32,16} = \frac{\|\mathbf{I}_{32} - \mathbf{I}_{16}\|_2}{\left\|\Delta\mathbf{Y}_{t,H}^{(PF)}\right\|_2 + 10^{-12}} \qquad \text{conv}_{16,8} = \frac{\|\mathbf{I}_{16} - \mathbf{I}_{8}\|_2}{\left\|\Delta\mathbf{Y}_{t,H}^{(PF)}\right\|_2 + 10^{-12}}$$

(F10)

The registered numerical-validity gates are applied to checkpoint-stratum medians over all 64 units: the median 32-subinterval endpoint-accounting error must be strictly $< 0.01$, and the median 32-versus-16 nested-grid convergence readback must be strictly $< 0.005$. No unit is removed based on its numerical-error magnitude before these medians are taken. The 8- and 16-subinterval endpoint-accounting errors and the 16-versus-8 convergence readback are stored diagnostics with no separate registered pass threshold.

These gates verify path accounting and are not additional scientific outcomes. Complete finite-operator and path-node coverage, correct array shapes, bound artifact identities, and exact original-dtype serialization/readback are separate integrity requirements. All 33 directional-response nodes and all three Simpson integrals are retained; full response operators are retained at hidden-dose values 0, 0.25, 0.5, 0.75, and 1.

### F.5 Interpretation boundary

The finite-dose assay separates two questions. $E_{\mathrm{local}}$ asks whether one factual-endpoint tangent is sufficient for the full hidden patch under the registered future query, whereas $C_R(0.5)$ asks whether the local full-carrier future-response operator departs from the straight endpoint chord at the registered midpoint. The path integral in Eq. (F7) is an established calculus identity and is used only to check that the sequence of local responses numerically accounts for the observed finite future change.

Accordingly, a finite $E_{\mathrm{local}}$ or $C_R(0.5)$ does not define an effective-state dimension, identify an intrinsic manifold curvature, or establish closure. These quantities characterize local-to-finite response behavior along the inherited factual-to-patched path in the registered $H = 8$ rollout setting. Formal thresholds, numerical-validity requirements, and checkpoint-first replication rules are specified in Section 3.7 and Appendix G.

# Appendix G. Confirmatory populations, aggregation, and frozen decision rules

## G.1 Certified response panel, frozen registries, and strata

The certified $H = 8$ response panel contains checkpoints 291402, 291403, and 291404 under the frozen contract used for Sections 3.3–3.7. The same identities had earlier exploratory use, so this assay is not treated as three additional independently trained checkpoints for the later tangent-transport study. Its support counts remain specific to the $H = 8$ response assay and are not added to the later fixed ten-checkpoint cohort.

The main response registry contains 128 matched units: 64 in S1 and 64 in S2. The same registry is evaluated under each of the three checkpoints for future-sensitive alignment and finite-dose reconfiguration. The causal-validation registry is a frozen 32-unit subset containing 16 units from each stratum. Each causal unit is evaluated separately at the patched and native-counterfactual base routes using 52 directions from the shared construction rules in Appendix E. Vector coordinates are route-specific apart from the four shared $\mathbf{U}_4$ directions. Unit identities, stratum assignments, subset membership, and population source are taken from the frozen registry rather than reconstructed from outcome tables.

No unit is added, removed, or reassigned from its observed response, and no direction or hidden-dose node is selected after seeing its effect. The checkpoint is the replication unit; lower-level repeats estimate only checkpoint-level summaries.

## G.2 Aggregation hierarchy

Aggregation is performed in a fixed order. First, every registered metric is computed at its native unit level. The alignment and finite-dose quantities contribute one value per matched unit. The perturbation assay evaluates 52 directions from the registered construction at each base route and reduces their predicted and measured effects to one Spearman rank correlation per unit and route. Second, unit-level values are summarized by the median within each checkpoint and dynamical stratum; causal predictiveness is additionally separated by patched and native-counterfactual route. Third, the frozen component classifier is applied to these checkpoint-stratum, or checkpoint-stratum-route, summaries. Only after this step are checkpoint support counts compared with the panel-level replication rule.

This hierarchy prevents the 52-direction bank, 33-node dose path, repeated routes, or within-checkpoint unit count from inflating the number of independent replications. They provide within-checkpoint precision and diagnostic resolution only, under the same aggregation order for supporting and failing cells.

## G.3 Frozen future-sensitive alignment rule

The formal alignment quantity is the CF-directed subspace gain defined in Section 3.4 and Appendix D. Response-assay contrastability is defined from the full-operator distance in Eq. (D6): a checkpoint-stratum cell is contrastable when the median over its 64 registered units is strictly greater than $1.807543226620821 \times 10^{-5}$. This numerical floor was frozen before the response assay and is distinct from the release-level restart contrastability condition in Eq. (B13).

All 64 unit-level alignment gains contribute to the cell median; units are not masked by their individual operator distances. A contrastable cell passes when its median alignment gain is strictly positive, and a checkpoint supports alignment only when both S1 and S2 are contrastable and positive. The classifier requires at least two contrastable checkpoint cells per stratum to be evaluable and at least two supportive checkpoints out of the fixed three for replication. For $m$ contrastable checkpoint summaries, the family summary is the sorted entry at zero-based index $\left\lfloor \frac{m}{2} \right\rfloor$: the usual median for $m = 3$ and the upper value for $m = 2$. A noncontrastable cell cannot support the component and does not reduce the fixed three-checkpoint denominator. All three checkpoints are contrastable in both strata in the accepted result, yielding 3/3 support. Supporting/descriptive operator distances cannot rescue a failed alignment gate.

### G.4 Frozen causal-predictiveness and finite-dose rules

For causal predictiveness, the unit-level statistic is $\rho^{(q)}$ from Section 3.5 and Appendix E. Within every checkpoint, stratum, and base route, the median of evaluable unit-level correlations must be strictly greater than 0.70. The patched and native-counterfactual routes must both pass in S1 and S2. A checkpoint supports the causal component only when all four route-by-stratum cells pass. Panel-level replication requires support from at least two of the three checkpoints. A constant predicted or measured rank vector is non-evaluable under the rule in Appendix E rather than silently assigned a passing correlation. The patched and native-counterfactual rank-agreement statistics are tested separately. The rule contains no threshold on their difference and no paired comparison of effects on identical cross-route vectors.

For finite-dose reconfiguration, the two registered effect quantities from Section 3.6 and Appendix F are summarized by the median over all 64 units in each checkpoint-stratum cell; both medians must be strictly > 0.10. Only the midpoint operator-chord deviation is the formal path-reconfiguration effect gate. Numerical validity additionally requires the two cell-median inequalities in Appendix F.4: 32-subinterval endpoint-accounting error < 0.01 and 32-versus-16 convergence < 0.005.

All six checkpoint-stratum cells must be numerically valid; otherwise the finite-dose component is non-evaluable. Numerical failures are not removed from the denominator or converted into scientific negative results. With numerical validity established, a checkpoint supports finite-dose reconfiguration only if both S1 and S2 pass both effect thresholds, and replication requires support from at least two of the fixed three checkpoints. All six cells are numerically valid in the accepted result. Checkpoints 291402 and 291403 support the component; checkpoint 291404 fails the S1 midpoint effect threshold (median 0.07963293432461849), leaving the reported 2/3 support unchanged.

### G.5 Non-rescue rule, replication scope, and provenance boundary

All decision rules are conjunctive at the level specified above. Failure of a required stratum, route, effect threshold, contrastability condition, or numerical-validity gate remains a failure for that checkpoint component. Supporting, descriptive, and post-hoc quantities cannot promote a failed primary component to support. The thresholds and route requirements are not retuned after inspecting the outcomes.

The checkpoint-first rule separates reproducibility across model instances from precision inside one model instance. A replicated response-assay component therefore means that the frozen criterion is satisfied by at least two checkpoint identities under all required strata and routes. It does not mean that every checkpoint behaves identically, that unit-, direction-, or dose-level measurements are independent model replications, or that the result is universal across architectures, environments, intervention families, or horizons.

Finally, the $H = 8$ response assay and the later tangent-transport confirmation answer distinct questions. The former tests future-response alignment, direction ranking, and finite-dose reconfiguration; the latter tests state-dependent transport and functional restart under its own horizon and cohorts. Their checkpoint counts are reported separately: the $H = 8$ panel is not additional tangent-transport replication and is not folded into the later fixed-sample denominators. Exact support counts and heterogeneity are reported in Results.

### G.6 Supporting and post-hoc reporting conventions

Supporting and post-hoc readbacks retain the frozen populations, route definitions, horizon, output metric, and checkpoint-first hierarchy of their parent analyses. The direct patched-factual subspace separation and full-operator readbacks contextualize the alignment result, while non-midpoint dose nodes describe progression along the fixed factual-to-patched hidden path.

These quantities do not define additional formal components, thresholds, or independent replications and cannot alter a registered component classification. They are reported only to characterize effect size and checkpoint- or stratum-specific heterogeneity. The fixed path remains a hidden factual-to-patched interpolation rather than a physical-edit dose axis or an interpolation toward native CF.

## Appendix H. Detailed result tables and checkpoint-level readbacks

This appendix reports checkpoint-level result tables and supporting readbacks for the main Results section.

### H.1 Structured-GRU transport/function confirmation and complete specificity

Across nine baseline-eligible structured-GRU checkpoints, all 18 checkpoint-by-stratum cells satisfied the transport-capture and functional criteria; median recovery spanned 0.744–1.034, with 17/18 cells above 0.85. The complete assay passed 16/18 eligible cells. Checkpoint 291405/S1 retained capture 0.938 and recovery 1.034 but failed contrastability and several specificity checks; checkpoint 291413/S2 retained capture 0.977 and recovery 0.965 but likewise failed contrastability/specificity. The original fresh panel supported the complete assay in 2/3 checkpoints and the fixed extension in 5/6 additional baseline-eligible checkpoints. Checkpoint 291414 failed baseline health. End-to-end support was therefore 7/10 (Wilson 95% interval 0.40–0.89), or 7/9 (0.45–0.94) among baseline-eligible checkpoints.

### H.2 Monolithic LSTM transport/function values

Table H1. Privileged tangent transport in fresh LSTM checkpoints 392001–392003. Entries are checkpoint–stratum summaries on the fixed 64-unit registry (32 S1 and 32 S2 units), using releases 1–10 and the aggregation conventions in Appendix B. All six cells pass the registered transport/function criteria.

| Checkpoint | S1 capture | S1 recovery | S2 capture | S2 recovery |
|---|---|---|---|---|
| 392001 | 0.975 | 0.942 | 0.973 | 0.983 |
| 392002 | 0.936 | 0.894 | 0.967 | 0.976 |
| 392003 | 0.997 | 0.986 | 0.971 | 0.959 |

All six cells passed the registered transport/function criteria. Contrastable coverage was 1.0 in every cell except 392001/S2 (0.95); positive-unit fraction was 1.0 throughout. Late-band transported-image capture ranged from 0.866 to 0.996 and late-band recovery from 0.809 to 0.977.

### H.3 Future-sensitive alignment and supporting readbacks

Table H2. Future-sensitive subspace alignment on the 128-unit response registry (64 units per stratum). Each entry is a checkpoint–stratum median. CF-directed gain is the primary contrast in Eq. (D5); positive values mean greater patched-to-native-CF than factual-to-native-CF overlap. P–F separation, Eq. (D4), is descriptive.

| Checkpoint | S1 CF-directed gain | S2 CF-directed gain | S1 P-F separation | S2 P-F separation |
|---|---|---|---|---|
| 291402 | 0.0920 | 0.0819 | 0.2350 | 0.0655 |
| 291403 | 0.0961 | 0.0975 | 0.0976 | 0.0965 |
| 291404 | 0.0043 | 0.0063 | 0.0167 | 0.0232 |

The fraction of units with positive CF-directed gain was 0.531 in 291404/S1 and 0.703 in 291404/S2, compared with 0.734–1.000 in the other four cells. The supporting CF-directed full-operator distance contrast was positive in five of six cells and slightly negative in 291404/S1 (-0.0064).

### H.4 Perturbation-predictiveness cell values

Table H3. Median unit-level Spearman correlations for the frozen 32-unit validation subset (16 units per stratum). Each unit is scored separately at patched (P) and native-counterfactual (CF) basepoints using 52 directions generated by the route-specific rules in Appendix E. Each entry summarizes one checkpoint–stratum–route cell, not a paired vectorwise route comparison.

| Checkpoint | S1 P | S1 CF | S2 P | S2 CF |
|---|---|---|---|---|
| 291402 | 0.999787 | 0.999744 | 0.999530 | 0.999616 |
| 291403 | 0.999658 | 0.999573 | 0.999616 | 0.999616 |
| 291404 | 0.999872 | 0.999872 | 0.999402 | 0.999530 |

Each unit-level correlation compares the response-operator prediction with independently measured finite autonomous effects on the same 52 route-specific registered directions. All 12 cells exceeded the preregistered threshold of 0.70.

### H.5 Finite-dose response and numerical path accounting

Table H4. Finite-dose response results on the 128-unit response registry (64 units per stratum). Entries are checkpoint–stratum medians of the normalized full-dose residual, Eq. (F3), and midpoint chord deviation, Eq. (F5). A cell passes only when both medians are strictly greater than 0.10 and numerical-validity checks pass.

| Checkpoint | Stratum | Full-dose residual | Midpoint chord deviation | Numerical valid | Cell |
|---|---|---|---|---|---|
| 291402 | S1 | 0.4187 | 0.2407 | Yes | Pass |
| 291402 | S2 | 1.1589 | 0.5202 | Yes | Pass |
| 291403 | S1 | 0.2480 | 0.1438 | Yes | Pass |
| 291403 | S2 | 0.8455 | 0.5124 | Yes | Pass |
| 291404 | S1 | 0.1888 | 0.0796 | Yes | Fail |
| 291404 | S2 | 0.6171 | 0.2790 | Yes | Pass |

In every checkpoint, both registered quantities were larger in S2 than in S1. The 32-subinterval endpoint-accounting error ranged from $1.26 \times 10^{-6}$ to $4.00 \times 10^{-6}$, and the 32-versus-16 subinterval convergence readback ranged from $1.25 \times 10^{-7}$ to $4.12 \times 10^{-5}$. These numerical readbacks confirm stable path accounting and are not separate scientific outcomes.